\documentclass[10pt,twocolumn,letterpaper]{article}

\usepackage[pagenumbers]{cvpr} 

\definecolor{cvprblue}{rgb}{0.21,0.49,0.74}
\usepackage[pagebackref,breaklinks,colorlinks,allcolors=cvprblue]{hyperref}

\usepackage{amsmath}
\usepackage{multirow}
\usepackage{adjustbox}
\usepackage{makecell}
\usepackage{wrapfig}

\usepackage[ruled,vlined]{algorithm2e}
\usepackage{graphicx}
\usepackage{colortbl}
\usepackage{array}
\usepackage{cuted}
\newcolumntype{C}{>{\centering\arraybackslash}p{1.5cm}}
\usepackage{makecell}
\newcommand{\best}[2]{\textbf{#1} $\pm$ #2}

\def\confName{CVPR}
\def\confYear{2027}

\title{Common Principal Component Networks for Domain Generalization}

\author{Yu-Hsi Chen\\
The University of Melbourne\\
{\tt\small yuhsi@student.unimelb.edu.au}
\and
Abd-Krim Seghouane\\
The University of Melbourne\\
{\tt\small abd-krim.seghouane@unimelb.edu.au}
}

\begin{document}
\maketitle
\begin{abstract}
Domain Generalization (DG) aims to learn representations that generalize to unseen target domains under out-of-distribution (OOD) shifts. Despite recent advances in invariant representation learning, explicitly capturing shared structures across source-domain distributions remains relatively underexplored. In this work, we integrate Common Principal Component Analysis (CPCA) into deep learning to identify a common orthogonal basis across multiple domains, proposing two complementary end-to-end differentiable frameworks. First, we introduce Common Principal Component Network (CPCNet), which unrolls the iterative Flury-Gautschi (FG) algorithm into fully differentiable neural layers, enabling covariance-based joint diagonalization within end-to-end representation learning. Second, to address rank deficiency arising from mini-batch covariance estimation in small-sample regimes, we develop Projection Pursuit-based Common Principal Component Network (PP-CPCNet), which directly optimizes a common orthogonal basis without explicit covariance estimation. Experiments validate the joint diagonalization capabilities of both methods, while evaluations on four standard DG benchmarks demonstrate competitive zero-shot generalization performance with favorable computational efficiency.
\end{abstract}    
\section{Introduction}
\label{sec:introduction}

\begin{figure}[!t]
    \centering
    \includegraphics[width=\linewidth]{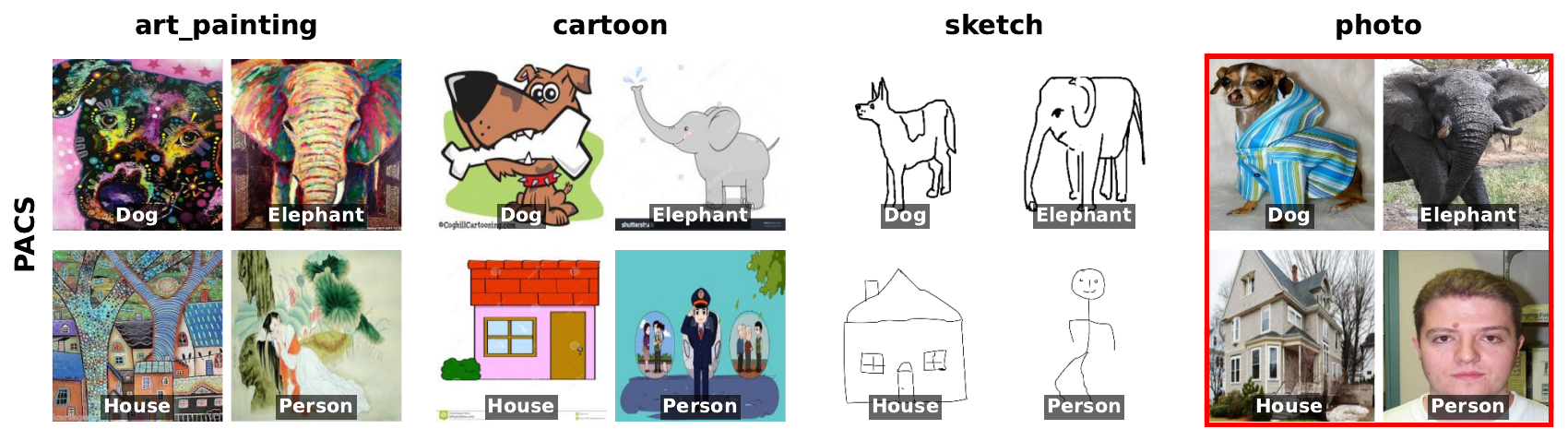}
    \caption{Illustration of the \texttt{PACS} dataset under the DG setting.}
    \label{fig:pacs_dataset}
\end{figure}

The Universal Approximation Theorem~\cite{hornik1989multilayer,cybenko1989approximation} establishes the expressive power of deep neural networks. However, their generalization performance typically relies on the assumption that training and test data are independently and identically distributed (i.i.d.)~\cite{bartlett2021deep}. In real-world applications, distribution shifts often violate this assumption, leading to substantial performance degradation~\cite{pan2009survey}. Domain Generalization (DG) addresses this challenge by learning robust and transferable representations from multiple source domains, with the goal of generalizing to unseen target domains~\cite{zhou2022domain,wang2022generalizing}. \Cref{fig:pacs_dataset} illustrates the DG setting using the \texttt{PACS} dataset.
Early DG methods primarily focused on reducing distribution discrepancies across source domains through statistical distance minimization~\cite{muandet2013domain,sun2016deep} and adversarial learning~\cite{ganin2016domain,li2018domain}. However, these approaches can struggle to capture complex semantic variations and handle the high dimensionality of visual data. More recent methods have therefore explored increasingly expressive architectures~\cite{tolstikhin2021mlp,touvron2022resmlp,hou2022vision} in conjunction with advanced regularization strategies~\cite{balaji2018metareg,kim2021selfreg,cha2022domain}.
Despite their empirical success, these methods often rely on increased model capacity to accommodate domain shifts rather than explicitly identifying shared structures across source domains.

A principled statistical approach to identifying shared structures across multiple distributions is Common Principal Component Analysis (CPCA)~\cite{flury1984common}. CPCA characterizes the second-order structure of multiple distributions through a shared orthogonal transformation that jointly diagonalizes their covariance matrices, thereby identifying common principal directions. However, conventional CPCA relies on the FG iterative algorithm~\cite{flury1986algorithm}, which is not directly amenable to end-to-end gradient-based optimization. Moreover, its linear formulation limits its ability to capture complex nonlinear structures in visual representations. These limitations have hindered the integration of CPCA into deep learning frameworks for DG. Motivated by these challenges, we investigate CPCA-based learning for DG. Our main contributions are summarized as follows:
\begin{itemize}
    \item \textbf{CPCA for Domain Generalization.} We formulate DG through CPCA, exploiting a shared orthogonal basis to characterize common structures across source domains.

    \item \textbf{CPCNet: Deep Unfolding of CPCA.} We develop a covariance-based architecture by unfolding the FG algorithm into differentiable layers, enabling end-to-end learning of the common orthogonal basis.

    \item \textbf{PP-CPCNet: Covariance-Free Formulation.} We propose a PP-based formulation that avoids mini-batch covariance estimation and mitigates rank deficiency.

    \item \textbf{Empirical Validation and Benchmarks.} We validate the joint diagonalization capabilities of both methods and evaluate their DG performance on four DomainBed benchmarks, demonstrating competitive zero-shot generalization and computational efficiency.
\end{itemize}
\section{Related Work}
\label{sec:related_work}

\subsection{Domain Generalization}
\label{ssec:rw_dg}

Domain Generalization (DG) addresses deep neural networks' vulnerability to out-of-distribution (OOD) shifts by learning representations that generalize reliably to unseen target domains. Unlike Domain Adaptation (DA)~\cite{ben2006analysis,you2019universal}, which typically assumes access to target-domain samples during training, DG operates in a zero-shot transfer setting without access to target-domain data.
Existing DG methods span a broad range of approaches, from foundational baselines such as Empirical Risk Minimization (ERM)~\cite{vapnik1998statistical} and its modern variants~\cite{teterwak2025erm++} to more specialized strategies for domain-invariant representation learning, including adversarial learning~\cite{yang2021adversarial,zhu2022localized,dayal2023madg}, causal learning~\cite{lv2022causality,jiang2022invariant}, feature disentanglement~\cite{zhang2022towards,wu2023uncovering,demirel2023adrmx}, and contrastive learning frameworks~\cite{kim2021selfreg,mahajan2021domain,jeon2021feature,verma2021towards,yao2022pcl}. Optimization-oriented approaches further enhance robustness through meta-learning-based domain shift simulation~\cite{li2018learning,balaji2018metareg,du2020learning,shu2021open}, gradient matching~\cite{shigradient}, and ensemble strategies~\cite{izmailov1803averaging,cha2021swad}. 

From an architectural perspective, DG has been explored across diverse backbone designs, including conventional CNNs~\cite{guo2023aloft}, MLP-based models~\cite{tolstikhin2021mlp,touvron2022resmlp,hou2022vision}, and Vision Transformers (ViTs)~\cite{sultana2022self,lisparse}. More recently, State Space Models (SSMs) have been investigated for efficient sequence modeling under domain shifts~\cite{liu2024vmamba,long2024dgmamba,guo2024start}, while pretrained vision-language models such as CLIP leverage shared image-text embedding spaces to transfer semantic knowledge across domains~\cite{zheng2026learning}.
Despite these advances, explicitly exploiting the shared statistical structure among source domains remains relatively underexplored. CPCA provides a principled approach by identifying a common basis that captures shared principal directions while allowing domain-specific variances to differ.

\subsection{Common Principal Component Analysis}
\label{ssec:rw_cpca}

Common Principal Component Analysis (CPCA)~\cite{flury1984common} is a multivariate statistical method that identifies a common orthogonal basis across multiple groups. Consider $K$ groups, where the $k$-th group consists of $N_k = n_k + 1$ independent samples, each represented by a $p$-dimensional random vector drawn from a multivariate normal distribution, $\mathcal{N}_p(\boldsymbol{\mu}_k, \boldsymbol{\Sigma}_k)$. Here, $\boldsymbol{\mu}_k \in \mathbb{R}^{p}$ and $\boldsymbol{\Sigma}_k \in \mathbb{R}^{p \times p}$ denote the population mean vector and positive definite covariance matrix, respectively. Let $\mathbf{S}_k$ denote the corresponding unbiased sample covariance matrix. Under the assumption that $n_k \ge p$ for all $k$, the sample covariance matrices follow independent Wishart distributions:
\begin{equation}
    n_k \mathbf{S}_k \sim \mathcal{W}_p(n_k, \boldsymbol{\Sigma}_k).
\end{equation}
CPCA assumes that the population covariance matrices $\{\boldsymbol{\Sigma}_k\}_{k=1}^K$ share a common basis $\boldsymbol{\beta} \in \mathbb{R}^{p \times p}$, such that
\begin{equation}
\label{eq:cpca_hypothesis}
    H_c:\:
    \boldsymbol{\beta}^{\top}\boldsymbol{\Sigma}_k\boldsymbol{\beta}
    = \boldsymbol{\Lambda}_k,
    \quad k = 1, \dots, K,
\end{equation}
where $\boldsymbol{\Lambda}_k = \mathrm{diag}(\lambda_{k1}, \ldots, \lambda_{kp})$ denotes a diagonal matrix containing the group-specific variances along the common principal components (CPCs). This formulation identifies shared principal directions across groups while allowing their corresponding variances to differ.

Under the maximum likelihood (ML) framework of~\cite{flury1984common}, the common basis $\boldsymbol{\beta}$ is estimated by enforcing the simultaneous diagonalization of the sample covariance matrices. The resulting stationarity conditions are expressed as
\begin{equation}
\label{eq:ml_statement}
    \boldsymbol{\beta}_{l}^{\top}
    \left(
    \sum_{k=1}^K n_k
    \frac{\lambda_{kl} - \lambda_{km}}
    {\lambda_{kl} \lambda_{km}}
    \mathbf{S}_{k}
    \right)
    \boldsymbol{\beta}_{m} = 0,
    \quad l \neq m,
\end{equation}
where $\boldsymbol{\beta}_l$ and $\boldsymbol{\beta}_m$ denote the $l$-th and $m$-th columns of $\boldsymbol{\beta}$, respectively, and $\lambda_{kl}$ and $\lambda_{km}$ are the corresponding diagonal entries of $\boldsymbol{\Lambda}_k$ in~\cref{eq:cpca_hypothesis}. The common basis is typically estimated using the iterative Flury-Gautschi (FG) algorithm~\cite{flury1986algorithm}. Given the estimated common basis, each group-specific data matrix $\mathbf{X}_k \in \mathbb{R}^{N_k \times p}$ is transformed as
\begin{equation}
\label{eq:sampled_cpcs}
    \mathbf{U}_k = \mathbf{X}_k \boldsymbol{\beta},
    \quad k = 1, \dots, K,
\end{equation}
where $\mathbf{U}_k$ contains the corresponding sample representations in the common principal component basis.
Building on its solid statistical foundation, CPCA has inspired a broad range of theoretical and methodological extensions, including partial CPCA under relaxed sharing assumptions~\cite{flury1987two}, robust estimators for handling outliers~\cite{boente2001robust,boente2002influence}, state-space formulations~\cite{gu2016raw}, and efficient stepwise optimization algorithms~\cite{trendafilov2010stepwise,riaz2021stepwise}. Beyond foundational studies~\cite{pepler2014identification}, CPCA has been extended to diverse statistical modeling settings~\cite{bagnato2021unconstrained,duras2022fixed,hu2023spectral} and applied to multiview representation learning~\cite{kanaan2018multiview} and multivariate time series clustering~\cite{li2019multivariate,ma2025fcpca,ma2026robcpca}.
Despite these advances, integrating CPCA into modern deep learning systems remains challenging due to its reliance on iterative joint diagonalization procedures, which involve non-differentiable operations.

\subsection{Deep Unfolding Neural Networks}
\label{ssec:rw_unfolding}

Deep Unfolding Neural Networks (DUNNs) bridge principled optimization and deep learning by transforming iterative algorithms into trainable neural architectures while retaining their underlying algorithmic structure and interpretability. This paradigm originated with Learned Iterative Shrinkage-Thresholding Algorithms (LISTA)~\cite{gregor2010learning}, which reformulated classical sparse coding solvers, such as ISTA and FISTA~\cite{daubechies2004iterative,rozell2008sparse,beck2009fast}, as learnable network layers. Subsequently, model-based constraints were incorporated into deep unfolding architectures for non-negative matrix factorization~\cite{hershey2014deep}.
Recent advances have further integrated deep unfolding with Transformer-based architectures~\cite{zhou2025dual,chen2026snapshot} and demonstrated its effectiveness across diverse applications, including wireless communications~\cite{balatsoukas2019deep,hu2020iterative,shi2022deep,feng2025deep,deka2026comprehensive} and computer vision tasks such as compressive sensing~\cite{zhang2018ista,you2021ista,song2021memory,song2023dynamic}, super-resolution~\cite{zhang2020deep,marivani2020multimodal,ma2021deep}, image restoration~\cite{kong2021deep,mou2022deep}, segmentation~\cite{wu2025rpcanet++,yang2026aw}, and small target detection~\cite{wu2024rpcanet,xiong2025drpca,liu2025ctvnet,liu2025ddfet,li2025small,liu2025lightweight,deng2025dusrnet,an2026du2stdnet}.
Building on this paradigm, we unfold the FG algorithm into a trainable architecture, incorporating the statistical structure of CPCA into an end-to-end deep learning framework for learning shared representations across domains.

\subsection{Projection Pursuit}
\label{ssec:rw_pp}

Projection Pursuit (PP)~\cite{friedman2006projection,huber1985projection} provides a statistical framework for identifying informative projection directions through the optimization of indices defined on projected observations. Unlike conventional variance maximization, PP allows different statistical criteria to guide projection learning, providing flexibility in characterizing the underlying data structure. In particular, robust projection indices based on $L_1$ dispersion~\cite{kwak2008principal} and median-based statistics~\cite{leys2013detecting} provide alternatives that reduce the influence of extreme observations, making learning less sensitive to outliers.
Beyond single-population analysis, PP has also been explored in CPCA. Unlike conventional CPCA, which estimates a common basis through the joint diagonalization of group-specific covariance matrices, projection-based approaches identify CPCs by optimizing statistical indices of projected observations. In particular, classical robust CPCA studies have developed PP-based estimators and established their robustness properties under suitable conditions~\cite{boente2001robust}, demonstrating the potential of projection-based formulations for robust common basis estimation.
Building on this statistical foundation, we develop PP-based common basis learning for deep neural networks for DG. This yields a covariance-free formulation of CPCA for learning shared representations across source domains.
\section{Methodology}
\label{sec:methodology}

In this section, we first formulate the use of CPCA for DG in~\cref{ssec:problem_formulation}, followed by two proposed end-to-end differentiable frameworks: covariance-based CPCNet in~\cref{ssec:cpcnet} and covariance-free PP-CPCNet in~\cref{ssec:ppcpcnet}.

\subsection{CPCA for Domain Generalization}
\label{ssec:problem_formulation}


Let $\mathcal{X} \subseteq \mathbb{R}^p$ denote the $p$-dimensional input space, and let $\mathcal{Y} = \{1, \dots, C\}$ denote the label space for a $C$-class classification task. A domain is characterized by a joint probability distribution $P_{XY}$ over $\mathcal{X} \times \mathcal{Y}$. In the standard DG setting, training data are available from $K$ distinct source environments, denoted by $\mathcal{E}_{tr} = \{E_k\}_{k=1}^{K}$, where $k$ indexes the source environments. Each source environment $E_k$ consists of $N_k$ samples, $\{(\mathbf{x}_i^{(k)}, y_i^{(k)})\}_{i=1}^{N_k}$, drawn from its corresponding joint distribution $P_{XY}^{(k)}$. The associated input samples are represented by a data matrix $\mathbf{X}_k \in \mathbb{R}^{N_k \times p}$. The fundamental objective of DG is to learn a robust predictive model using only the available source data, $\mathcal{E}_{tr}$, that minimizes the expected risk on a strictly unseen target environment $E_{te}$, characterized by a joint distribution $P_{XY}^{(te)}$ that differs from those of the source environments.

Let $h_\theta: \mathbb{R}^p \rightarrow \mathbb{R}^D$ denote a pre-trained neural backbone parameterized by $\theta$, which extracts high-dimensional features $\mathbf{f}_i^{(k)} = h_\theta(\mathbf{x}_i^{(k)})$. A projection module $b_\psi: \mathbb{R}^D \rightarrow \mathbb{R}^d$, parameterized by $\psi$, then maps these features into a lower-dimensional latent space, yielding $\mathbf{z}_i^{(k)} = b_\psi(\mathbf{f}_i^{(k)}) \in \mathbb{R}^d$ for the $i$-th sample in the $k$-th environment. Unlike classical CPCA, which operates directly on the original $p$-dimensional input space, we perform CPCA in this $d$-dimensional latent space. Throughout the paper, $\mathbf{S}_k$, $\boldsymbol{\beta}$, and $\mathbf{U}_k$ denote the covariance matrices, common basis, and domain-specific diagonal matrices in the latent space, respectively. To characterize the distributional structure of the latent features in each environment, we compute the unbiased sample covariance matrix $\mathbf{S}_k \in \mathbb{R}^{d \times d}$ as
\begin{equation}
\label{eq:cpcnet_Sk}
    \mathbf{S}_k = \frac{1}{n_k} \sum_{i=1}^{N_k} \left(\mathbf{z}_{i}^{(k)} - \bar{\mathbf{z}}^{(k)}\right)\left(\mathbf{z}_{i}^{(k)} - \bar{\mathbf{z}}^{(k)}\right)^\top,
\end{equation}
where $\bar{\mathbf{z}}^{(k)} \in \mathbb{R}^d$ denotes the sample mean vector of the latent features in the $k$-th environment, and $n_k = N_k - 1$.

We then solve the ML formulation in~\cref{eq:ml_statement} to obtain the common basis $\boldsymbol{\beta} \in \mathbb{R}^{d \times d}$ that simultaneously diagonalizes the $K$ latent source covariance matrices $\{\mathbf{S}_k\}_{k=1}^K$.
Let $\mathbf{Z}_k \in \mathbb{R}^{N_k \times d}$ denote the feature matrix whose rows correspond to the latent feature vectors $\mathbf{z}_i^{(k)}$. We project the latent representations onto the learned common basis to obtain the transformed feature matrix:
\begin{equation}
    \mathbf{U}_k = \mathbf{Z}_k \boldsymbol{\beta}, \quad k = 1, \dots, K.
\end{equation}
This transformation provides a shared coordinate system across source domains, aligning their covariance structures while preserving the underlying feature dimensionality.

During inference on an unseen target environment $E_{te}$, a target sample $\mathbf{x}^{(te)} \in \mathbb{R}^p$ is first mapped to a high-dimensional feature vector $\mathbf{f}^{(te)} = h_\theta(\mathbf{x}^{(te)}) \in \mathbb{R}^{D}$ and then transformed into a latent representation $\mathbf{z}^{(te)} = b_\psi(\mathbf{f}^{(te)}) \in \mathbb{R}^{d}$. The latent representation is subsequently transformed into the learned common coordinate system using the CPCA basis \(\mathbf{u}^{(te)} = \boldsymbol{\beta}^\top \mathbf{z}^{(te)}\). A straightforward inference strategy is to perform classification directly on the transformed representation. Specifically, the predicted class label $\hat{y}$ can be obtained using a linear classifier parameterized by $\mathbf{W}_{\text{cls}} \in \mathbb{R}^{d \times C}$ and $\mathbf{b}_{\text{cls}} \in \mathbb{R}^{C}$:
\begin{equation}
\label{eq:classifier}
    \hat{y} = \arg\max_{c \in \{1, \dots, C\}}
    \left( \mathbf{W}_{\text{cls}}^\top \mathbf{u}^{(te)} + \mathbf{b}_{\text{cls}} \right)_c.
\end{equation}
While the above formulation provides a principled framework for incorporating CPCA into DG, it presents two key challenges. First, standard CPCA estimation relies on iterative joint diagonalization, which has a computational complexity of $\mathcal{O}(d^3)$ per complete sweep, making it costly in high-dimensional feature spaces. Second, the FG algorithm employs a discrete swapping mechanism to select and order pairwise rotations, introducing non-differentiable operations that prevent its direct integration into end-to-end gradient-based optimization.

\subsection{CPCNet: Deep Unfolding of CPCA}
\label{ssec:cpcnet}

The CPCNet framework enables end-to-end differentiable optimization of CPCA through deep unfolding. Specifically, we first incorporate orthogonal retraction to maintain the orthogonality constraint (\cref{sssec:orthogonal_cayley}), then compute Riemannian gradients to enable gradient-based updates on the Stiefel manifold (\cref{sssec:riemannian_gradient}). We further introduce hypernetwork-based dynamic unfolding to adaptively determine the step sizes at each iteration (\cref{sssec:dynamic_unfolding}). Finally, we present the training objective (\cref{sssec:cpcnet_objective}) and the inference procedure (\cref{sssec:mgfm}). The complete CPCNet algorithm is provided in the supplementary material (\cref{alg:cpcnet}).

\subsubsection{Orthogonal Retraction via the Cayley Transform}
\label{sssec:orthogonal_cayley}

To maintain the orthogonality constraint on the shared basis $\boldsymbol{\beta}$ during gradient-based optimization, we employ the Cayley transform as a retraction on the orthogonal group~\cite{absil2008optimization}. The tangent space of the orthogonal group $\mathcal{O}(d)$ is characterized by the Lie algebra of skew-symmetric matrices, defined as $\mathfrak{so}(d)=\{\mathbf{A}\in\mathbb{R}^{d\times d}\mid \mathbf{A}^\top=-\mathbf{A}\}$. 
Rather than relying on computationally expensive orthogonalization procedures under the strict constraint $\boldsymbol{\beta}^\top \boldsymbol{\beta}=\mathbf{I}_d$, we parameterize its updates using an unconstrained vector $\boldsymbol{\theta}\in\mathbb{R}^{d(d-1)/2}$. A deterministic mapping $\mathcal{S}$ constructs the skew-symmetric matrix $\mathbf{A}=\mathcal{S}(\boldsymbol{\theta})$. We then map this to the orthogonal group through the Cayley transform~\cite{lezcano2019cheap,liefficient}:
\begin{equation}
    \boldsymbol{\beta}(\mathbf{A})=
    \left(\mathbf{I}_d-\frac{1}{2}\mathbf{A}\right)
    \left(\mathbf{I}_d+\frac{1}{2}\mathbf{A}\right)^{-1}.
    \label{eq:cayley_transform}
\end{equation}
This parameterization preserves orthogonality by construction, ensuring stable optimization.

\subsubsection{Riemannian Gradient Formulation}
\label{sssec:riemannian_gradient}

To enable gradient-based optimization on the orthogonal manifold, we derive the Riemannian gradient of the CPCA objective with respect to the skew-symmetric tangent space. Given an orthogonal basis $\boldsymbol{\beta}$, the basis-transformed variances are defined as the diagonal entries of the transformed covariance matrices, namely $\lambda_{kl} = [\boldsymbol{\beta}^\top \mathbf{S}_k \boldsymbol{\beta}]_{l,l}$. Following~\cite{flury1984common}, the ML estimate of the common basis $\boldsymbol{\beta}$ is obtained by minimizing the negative log-likelihood:
\begin{equation}
\label{eq:flury_objective}
    \mathcal{J}(\boldsymbol{\beta})
    =
    \sum_{k=1}^K n_k \sum_{l=1}^d \log(\lambda_{kl}).
\end{equation}
Taking the derivative of this objective with respect to $\boldsymbol{\beta}$ yields the Euclidean gradient:
\begin{equation}
\label{eq:G_euc}
    \mathbf{G}_{\text{Euc}} =
    \frac{\partial \mathcal{J}(\boldsymbol{\beta})}
    {\partial \boldsymbol{\beta}}
    =
    2 \sum_{k=1}^K n_k
    \mathbf{S}_k \boldsymbol{\beta} \boldsymbol{\Lambda}_k^{-1},
\end{equation}
where $\boldsymbol{\Lambda}_k = \operatorname{diag}(\lambda_{k1}, \dots, \lambda_{kd})$.

Since $\boldsymbol{\beta}$ lies on the Stiefel manifold $St(d,d)$, which coincides with the orthogonal group $\mathcal{O}(d)$, unconstrained Euclidean updates may violate the orthogonality constraint. Following the canonical metric geometry of the orthogonal group~\cite{edelman1998geometry}, we instead project the Euclidean gradient onto the tangent space $\mathfrak{so}(d)$. Specifically, following Cayley-based optimization frameworks~\cite{wen2013feasible}, the corresponding gradient representation is given by:
\begin{equation}
    \mathbf{G}_{\mathbf{A}} =
    \boldsymbol{\beta}^\top \mathbf{G}_{\text{Euc}}
    -
    \mathbf{G}_{\text{Euc}}^\top \boldsymbol{\beta}.
\end{equation}
Expanding this expression element-wise yields:
\begin{equation}
    [\mathbf{G}_{\mathbf{A}}]_{l,m}
    =
    2 \sum_{k=1}^K n_k
    [\boldsymbol{\beta}^\top \mathbf{S}_k \boldsymbol{\beta}]_{l,m}
    \left(
    \frac{\lambda_{kl}-\lambda_{km}}
    {\lambda_{kl}\lambda_{km}}
    \right).
\end{equation}

To facilitate efficient implementation through batched tensor operations within the computational graph, we introduce a skew-symmetric weight matrix $\boldsymbol{\Omega}_k \in \mathbb{R}^{d\times d}$ for each domain, whose entries encode pairwise variance differences. For numerical stability, particularly when mini-batch variances approach zero, we introduce a small constant $\epsilon$ in the denominator. Absorbing the scalar factor $2$ into the learning rate, we define:
\begin{equation}
    [\boldsymbol{\Omega}_k]_{l,m}
    =
    \frac{\lambda_{kl}-\lambda_{km}}
    {\lambda_{kl}\lambda_{km}+\epsilon}.
\end{equation}
The skew-symmetric gradient can then be efficiently computed using the Hadamard (element-wise) product $\odot$:
\begin{equation}
    \mathbf{G}_{\mathbf{A}}
    =
    \sum_{k=1}^{K} n_k
    \left(
    (\boldsymbol{\beta}^\top \mathbf{S}_k \boldsymbol{\beta})
    \odot \boldsymbol{\Omega}_k
    \right).
\end{equation}
Finally, we normalize the gradient by its Frobenius norm to stabilize gradient updates during deep unfolding.

\subsubsection{Dynamic Unfolding via Hypernetworks}
\label{sssec:dynamic_unfolding}

We unfold Riemannian gradient descent into $T$ differentiable layers. In conventional algorithm unrolling, step sizes are typically fixed or learned as static parameters. However, under DG, covariance statistics vary across mini-batches, making static step sizes potentially suboptimal. Inspired by dynamic parameter generation~\cite{xiong2025drpca}, we adopt a lightweight hypernetwork $H_\phi$ that takes the flattened covariance matrices as input and generates a context-dependent step-size vector $\boldsymbol{\eta}\in\mathbb{R}^T$ for the unfolded layers:
\begin{equation}
    \boldsymbol{\eta}
    =
    \frac{1}{2}\sigma\big(H_\phi(\mathbf{S}_1,\dots,\mathbf{S}_K)\big),
\end{equation}
where $\sigma$ denotes the sigmoid function. This parameterization constrains each step size to $(0,0.5)$, providing bounded and adaptive updates in response to mini-batch variations.

Starting from the origin of the Lie algebra ($\mathbf{A}_0=\mathbf{0}$), the network iteratively accumulates skew-symmetric updates over $T$ unfolded layers. At each layer $t=1,\dots,T$, the normalized Riemannian gradient $\tilde{\mathbf{G}}_{\mathbf{A},t-1}$ is used to update the skew-symmetric matrix $\mathbf{A}_t$, followed by the Cayley transform to obtain the corresponding orthogonal basis $\boldsymbol{\beta}_t$:
\begin{equation}
\resizebox{\linewidth}{!}{$
    \mathbf{A}_t
    =
    \mathbf{A}_{t-1}
    -
    \eta_t\tilde{\mathbf{G}}_{\mathbf{A},t-1},
    \quad
    \boldsymbol{\beta}_t
    =
    \left(\mathbf{I}_d-\frac{1}{2}\mathbf{A}_t\right)
    \left(\mathbf{I}_d+\frac{1}{2}\mathbf{A}_t\right)^{-1}.
$}
\end{equation}
The final basis $\boldsymbol{\beta}_T$ is orthogonal and differentiable, adapting to the covariance statistics of each forward pass.

\subsubsection{The CPCA Regularization Objective}
\label{sssec:cpcnet_objective}

While the unfolded module dynamically estimates the basis $\boldsymbol{\beta}_T$ for each batch, the backbone must learn representations amenable to joint diagonalization. To enforce this structural prior, we penalize the off-diagonal energy of the basis-transformed covariances. Let $\hat{\mathbf{S}}_k = \boldsymbol{\beta}_T^\top \mathbf{S}_k \boldsymbol{\beta}_T$ denote the covariance matrix in the learned basis for the $k$-th domain. The CPC regularization is defined as:
\begin{equation}
    \mathcal{L}_{\text{cpc}} =
    \frac{1}{K} \sum_{k=1}^K
    \frac{\|\hat{\mathbf{S}}_k\|_F^2 - \|\mathrm{diag}(\hat{\mathbf{S}}_k)\|_F^2}{d(d-1)}.
\end{equation}
The full training objective is then given by:
\begin{equation}
    \mathcal{L}_{\text{total}} = \alpha_{\text{task}}\mathcal{L}_{\text{task}} + \alpha_{\text{cpc}} \mathcal{L}_{\text{cpc}},
\end{equation}
where $\mathcal{L}_{\text{task}}$ denotes the cross-entropy loss for DG, and $\alpha_{\text{task}}$ and $\alpha_{\text{cpc}}$ weight the task and CPC objectives, respectively.

\subsubsection{Manifold-Guided Feature Modulation}
\label{sssec:mgfm}

The learned orthogonal basis $\boldsymbol{\beta}_T$ captures the shared geometric structure. Rather than directly classifying the low-dimensional representation, as formulated in~\cref{ssec:problem_formulation}, we use it to guide the modulation of the high-dimensional backbone features. Specifically, let $\mathbf{f} \in \mathbb{R}^{D}$ denote the backbone features and $\mathbf{z} \in \mathbb{R}^{d}$ the corresponding low-dimensional representation, where $D \gg d$. We obtain the invariant representation by projecting $\mathbf{z}$ onto the learned basis:
\begin{equation}
    \mathbf{u} = \boldsymbol{\beta}_T^\top \mathbf{z}.
\end{equation}
The representation $\mathbf{u}$ is then fed into two lightweight MLPs to generate channel-wise scaling and shifting vectors:
\begin{equation}
    \boldsymbol{\gamma} = 2 \sigma\big(\mathrm{MLP}_1(\mathbf{u})\big), \quad
    \Delta \mathbf{f} = \mathrm{MLP}_2(\mathbf{u}),
\end{equation}
where $\sigma$ denotes the sigmoid function and $\boldsymbol{\gamma}, \Delta \mathbf{f} \in \mathbb{R}^{D}$. To preserve the pre-trained backbone at initialization, we zero-initialize the weights and biases of the output layers of $\mathrm{MLP}_1$ and $\mathrm{MLP}_2$, yielding $\boldsymbol{\gamma} = \mathbf{1}$ and $\Delta \mathbf{f} = \mathbf{0}$. This initializes the model as the standard ERM baseline~\cite{vapnik1998statistical}, allowing the proposed modulation to be learned progressively during training.
Following Feature-wise Linear Modulation (FiLM)~\cite{perez2018film,turkoglu2022film}, we apply a channel-wise affine modulation to the backbone features:
\begin{equation}
    \tilde{\mathbf{f}} = \mathbf{f} \odot \boldsymbol{\gamma} + \Delta \mathbf{f},
\end{equation}
where $\odot$ denotes the Hadamard product. Analogously to~\cref{eq:classifier}, we employ a linear classifier with weight matrix $\mathbf{W}_{\text{cls}} \in \mathbb{R}^{D \times C}$ and bias vector $\mathbf{b}_{\text{cls}} \in \mathbb{R}^{C}$:
\begin{equation}
    \hat{y} = \arg\max_{c \in \{1,\dots,C\}}
    \left(\mathbf{W}_{\text{cls}}^{\top}\tilde{\mathbf{f}}+\mathbf{b}_{\text{cls}}\right)_c.
\end{equation}

\subsection{PP-CPCNet: Covariance-Free Formulation}
\label{ssec:ppcpcnet}

The PP-CPCNet framework addresses the mini-batch singularity problem in covariance-based optimization, which arises when the number of mini-batch samples is insufficient relative to the latent feature dimension, through PP. Specifically, we introduce parametric Stiefel manifold optimization for stable basis learning without explicit mini-batch covariance estimation (\cref{sssec:parametric_stiefel}), followed by a PP index that identifies informative projection directions and breaks the symmetry of the optimization objective (\cref{sssec:symmetry_breaking}). We then present the PP-CPCNet architecture (\cref{sssec:pbc}) and its training objective and inference procedure (\cref{sssec:ofi}). The complete PP-CPCNet algorithm is provided in the supplementary material (\cref{alg:pp_cpcnet}).

\subsubsection{Parametric Stiefel Manifold Optimization}
\label{sssec:parametric_stiefel}

Let $\tilde{\mathbf{F}}_{t-1}\in\mathbb{R}^{B\times D}$ denote the input features at cascade depth $t$, where $B=\sum_{k=1}^K N_k$ is the total mini-batch size. A bottleneck module $b_{\psi_t}$ maps these features to $\mathbf{Z}_t\in\mathbb{R}^{B\times d_t}$, where $d_t$ is the projection dimension. Following the orthogonal parameterization in~\cref{sssec:orthogonal_cayley}, we represent the depth-specific basis $\boldsymbol{\beta}_t$ using an unconstrained vector $\boldsymbol{\theta}_t\in\mathbb{R}^{d_t(d_t-1)/2}$. This vector is mapped to a skew-symmetric matrix $\mathbf{A}_t=\mathcal{S}(\boldsymbol{\theta}_t)$, from which the Cayley transform defined in~\cref{eq:cayley_transform} yields the orthogonal basis $\boldsymbol{\beta}_t$. The projected features are then given by
\begin{equation}
    \mathbf{U}_t=\mathbf{Z}_t\boldsymbol{\beta}_t.
\end{equation}

\subsubsection{Symmetry-Breaking Robust Dispersion}
\label{sssec:symmetry_breaking}

PP identifies informative components by maximizing the dispersion of projected features $\mathbf{U}_t$. However, since $\boldsymbol{\beta}_t$ is orthogonal, the total variance of $\mathbf{U}_t$ is invariant under rotation~\cite{richman1986rotation}, so an unweighted variance objective cannot distinguish between orthogonal bases. To break this rotational symmetry and promote variance concentration along leading components, we introduce a monotonically decreasing weight vector $\mathbf{w}_t\in\mathbb{R}^{d_t}$, whose entries are proportional to $[d_t,d_t-1,\ldots,1]$ and normalized to sum to one. Under trace invariance, this weighting drives orthogonal bases that concentrate greater dispersion along leading axes, thereby encouraging alignment with dominant eigenvectors and suppressing off-diagonal covariance components. By aggregating weighted dispersion across all $K$ domains, the resulting objective promotes a common orthogonal basis that jointly diagonalizes their feature covariances without explicitly constructing covariance matrices.

To improve robustness against domain-shifted outliers, we replace standard $L_2$ variance, which is sensitive to extreme values, with a detached-median $L_1$ dispersion metric. Although the Median Absolute Deviation (MAD) is robust to outliers~\cite{leys2013detecting}, its non-smooth median operator can produce sparse gradients. Our formulation retains robust dispersion estimation while providing denser optimization signals. Specifically, for domain $k$ at cascade depth $t$, let $\mathbf{U}_{t,k}^{(i)}$ denote the projected feature of the $i$-th sample. The median anchor is defined as
\begin{equation}
    \mathbf{m}_{t,k}
    =
    \mathrm{sg}\!\left[
    \mathrm{median} \big( \{ \mathbf{U}_{t,k}^{(i)} \}_{i=1}^{N_k} \big)
    \right],
\end{equation}
where $\mathrm{sg}[\cdot]$ denotes the stop-gradient operator, preventing gradients from propagating through the median computation~\cite{chen2021exploring}. The resulting PP dispersion score is
\begin{equation}
    s_{t,k}
    =
    \mathbf{w}_t^\top
    \left(
    \frac{1}{N_k}
    \sum_{i=1}^{N_k}
    \left|
    \mathbf{U}_{t,k}^{(i)}
    -
    \mathbf{m}_{t,k}
    \right|
    \right).
\end{equation}

\subsubsection{Progressive Bottleneck Cascade}
\label{sssec:pbc}

In PP-CPCNet, we employ a progressive bottleneck cascade consisting of $T$ layers. Inspired by the information bottleneck principle~\cite{tishby2000information,hinton2006reducing}, the bottleneck dimension progressively decreases across cascade depths, satisfying $d_1 > d_2 > \dots > d_T$. Specifically, the dimension is halved at each depth, subject to a minimum of 16:
\begin{equation}
    d_{t+1} = \max\left(\left\lfloor\frac{d_t}{2}\right\rfloor, 16\right).
\end{equation}
This progressive dimensionality reduction encourages the network to suppress domain-specific variations while retaining increasingly compact semantic representations.

\subsubsection{Objective Formulation and Inference}
\label{sssec:ofi}

We maximize PP dispersion by minimizing the negative sum of the dispersion scores, normalized by $KT$ to account for the number of source domains and cascade depths:
\begin{equation}
    \mathcal{L}_{\text{ppcpc}}
    =
    -\frac{1}{K T}
    \sum_{t=1}^T \sum_{k=1}^K s_{t,k}.
\end{equation}
The overall objective is then given by:
\begin{equation}
    \mathcal{L}_{\text{total}}
    =
    \alpha_{\text{task}}\mathcal{L}_{\text{task}}
    +
    \alpha_{\text{ppcpc}}\mathcal{L}_{\text{ppcpc}},
\end{equation}
where $\mathcal{L}_{\text{task}}$ denotes the cross-entropy loss for DG, and $\alpha_{\text{task}}$ and $\alpha_{\text{ppcpc}}$ weight the contributions of the two objectives.

During training and inference, we apply the feature modulation described in~\cref{sssec:mgfm} at each cascade depth to refine the backbone feature representation. For an input sample, the modulated feature at cascade depth $t$ is obtained as:
\begin{equation}
    \tilde{\mathbf{f}}_t
    =
    \left(\tilde{\mathbf{f}}_{t-1}\odot\boldsymbol{\gamma}_t\right)
    +\Delta\mathbf{f}_t,
\end{equation}
where $\boldsymbol{\gamma}_t\in\mathbb{R}^{D}$ and $\Delta\mathbf{f}_t\in\mathbb{R}^{D}$ denote the feature-wise scaling and residual adjustment, respectively, and $\odot$ denotes the Hadamard product.
At the final cascade depth, we use the final modulated feature representation $\tilde{\mathbf{f}}_T\in\mathbb{R}^{D}$ as input to a linear classifier, with weight matrix $\mathbf{W}_{\text{cls}}\in\mathbb{R}^{D\times C}$ and bias vector $\mathbf{b}_{\text{cls}}\in\mathbb{R}^{C}$, to predict the class label:
\begin{equation}
    \hat{y}
    =
    \arg\max_{c\in\{1,\ldots,C\}}
    \left(
    \mathbf{W}_{\text{cls}}^\top\tilde{\mathbf{f}}_T
    +\mathbf{b}_{\text{cls}}
    \right)_c.
\end{equation}
\section{Experiments}
\label{sec:experimental_results}

In this section, we first validate the joint diagonalization capabilities of CPCNet and PP-CPCNet (\cref{ssec:validation_of_cpcs}). We then present the implementation details and main results on DG benchmarks (\cref{ssec:DG_benchmarks}), and conclude with ablation studies and further analysis (\cref{ssec:ablation_analysis}).

\subsection{Validation of Common Principal Components}
\label{ssec:validation_of_cpcs}

We first validate CPCNet and PP-CPCNet's joint diagonalization capabilities on the Iris dataset~\cite{iris_53} ($K=3$, $d=4$), independent of downstream classification. To isolate their alignment objectives, we disable task-specific gradients by setting $\alpha_{\text{task}}=0$, with $\alpha_{\text{cpca}}=1$ for CPCNet and $\alpha_{\text{ppcpc}}=1$ for PP-CPCNet. We compare the optimization trajectories against the standard CPCA baseline using two criteria. Let $\hat{\mathbf{S}}_k = \boldsymbol{\beta}^\top \mathbf{S}_k \boldsymbol{\beta}$ denote the covariance matrix of domain $k$ transformed by the common orthogonal basis $\boldsymbol{\beta}$. The first is the ML objective~\cite{flury1984common}, which measures the deviation from joint diagonalization through the log-ratio of the product of diagonal entries to the determinant:
\begin{equation}
    \mathcal{M}_{\text{ML}} = \sum_{k=1}^K n_k \ln \left( \frac{\prod_{j=1}^d [\hat{\mathbf{S}}_k]_{jj}}{\det(\hat{\mathbf{S}}_k)} \right).
\end{equation}
By Hadamard's inequality, $\mathcal{M}_{\text{ML}}\geq 0$, with equality if and only if $\{\hat{\mathbf{S}}_k\}_{k=1}^K$ are diagonal. Thus, lower values indicate a better fit to the common principal components model. The second is the least-squares (LS) criterion, which measures the weighted off-diagonal Frobenius energy:
\begin{equation}
    \mathcal{M}_{\text{LS}}
    =
    \sum_{k=1}^K n_k
    \left(
    \|\hat{\mathbf{S}}_k\|_F^2
    -
    \|\mathrm{diag}(\hat{\mathbf{S}}_k)\|_F^2
    \right).
\end{equation}
Lower values indicate better joint diagonalization, with $\mathcal{M}_{\text{LS}}=0$ if and only if $\{\hat{\mathbf{S}}_k\}_{k=1}^K$ are diagonal.

\begin{figure}[!t]
    \centering
    \includegraphics[width=0.71\linewidth]{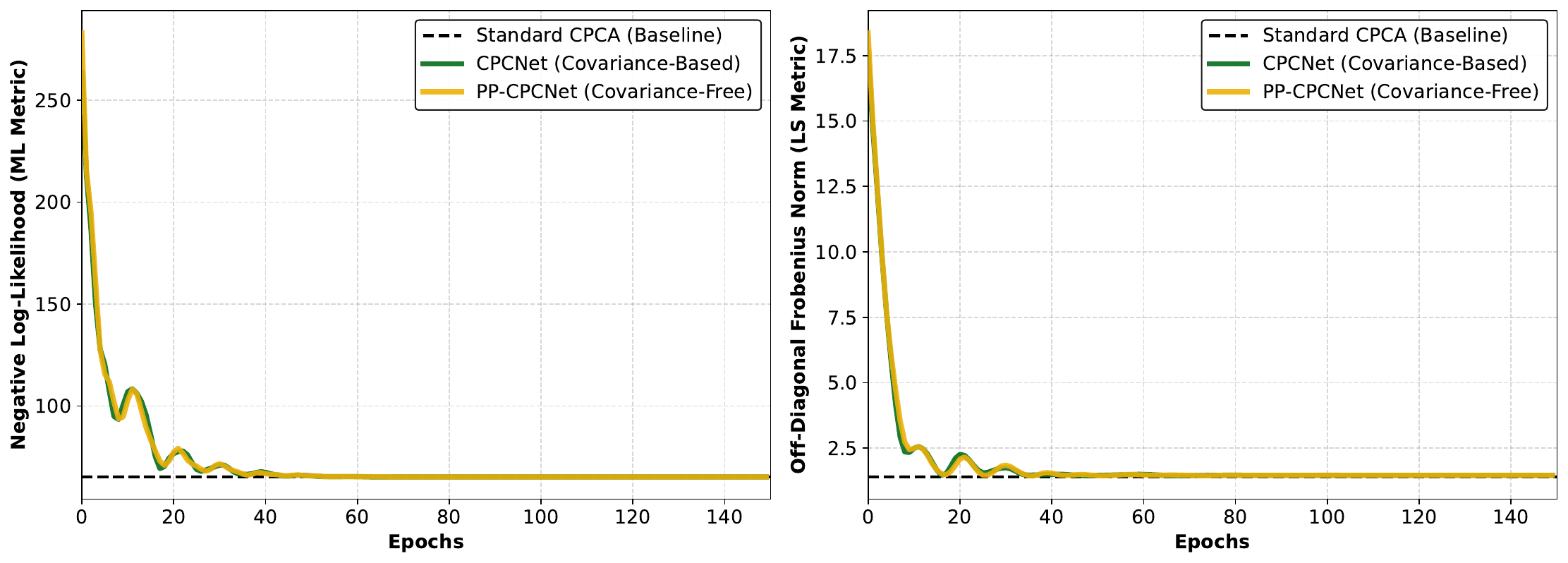}
    \caption{Convergence trajectories of joint diagonalization.}
    \label{fig:iris_trajectory}
\end{figure}

\begin{figure}[!t]
    \centering
    \includegraphics[width=0.71\linewidth]{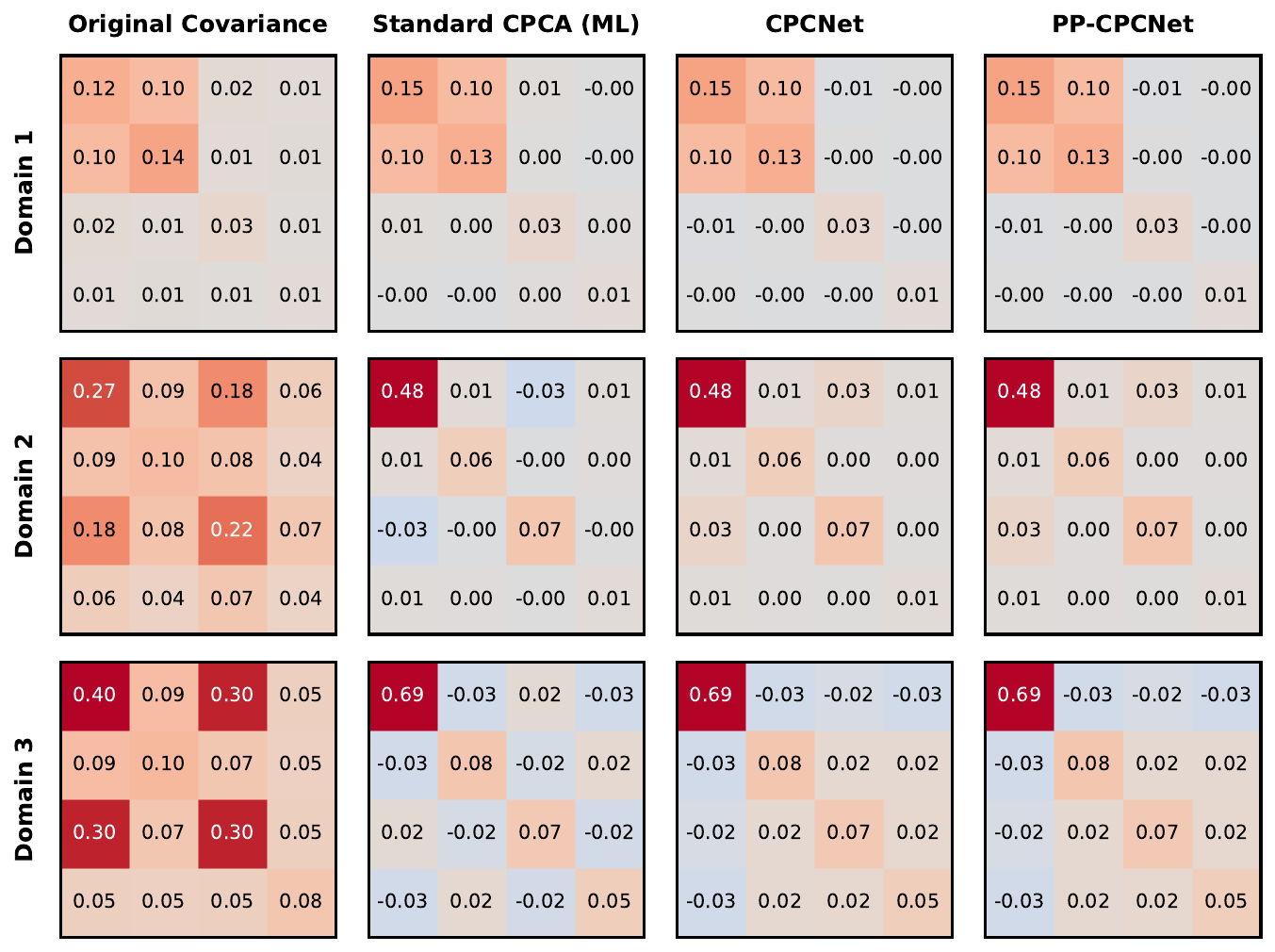}
    \caption{Visualization of joint diagonalization.}
    \label{fig:iris_matrices}
\end{figure}

As reported in~\cref{tab:iris_metrics}, both CPCNet and PP-CPCNet achieve the same ML objective as the standard CPCA baseline, with stable convergence after 118 and 121 epochs, respectively, as shown in~\cref{fig:iris_trajectory}. The highly overlapping loss curves further confirm that PP-CPCNet recovers the common orthogonal basis without observable optimization degradation in the full-sample regime, where covariance estimates are well-conditioned, despite bypassing explicit covariance construction.
This agreement is further illustrated in~\cref{fig:iris_matrices}, where both proposed methods effectively suppress off-diagonal covariance components across all three domains. These results demonstrate that both formulations can recover the joint diagonalization achieved by standard CPCA through end-to-end differentiable optimization.

\begin{table}[t]
    \centering
    \caption{Quantitative comparison of joint diagonalization.}
    \label{tab:iris_metrics}
    \resizebox{\linewidth}{!}{
    \begin{tabular}{lccc}
        \toprule
        \textbf{Method} & \textbf{Epochs} & \textbf{ML Metric ($\downarrow$)} & \textbf{LS Metric ($\downarrow$)} \\
        \midrule
        Standard CPCA (ML Baseline) & -- & \textbf{65.2142} & 1.4641 \\
        Standard CPCA (LS Baseline) & -- & 73.5022 & \textbf{1.4007} \\
        \midrule
        CPCNet (Covariance-Based) & 118 & \textbf{65.2142} & 1.4642 \\
        PP-CPCNet (Covariance-Free) & 121 & \textbf{65.2142} & 1.4643 \\
        \bottomrule
    \end{tabular}
    }
\end{table}

\subsection{Domain Generalization Benchmarks}
\label{ssec:DG_benchmarks}

We evaluate CPCNet and PP-CPCNet on four widely used DG benchmarks: \texttt{PACS}, \texttt{VLCS}, \texttt{OfficeHome}, and \texttt{TerraIncognita}. Detailed dataset statistics and download sources are provided in the supplementary material (\cref{tab:detailed_dataset_summary}) to clarify inconsistencies in prior literature, including domain names and image counts, and facilitate reproducibility and fair comparisons.
Training configurations in prior DG studies vary substantially in training duration, batch size, and optimizer settings, complicating fair comparisons across methods. We therefore adopt the unified training and validation protocol of DomainBed~\cite{gulrajanisearch}, following the hyperparameter configurations of~\cite{cha2021swad}, which also mentioned the benefits of extended training. All models use a batch size of 32 per domain, a dropout rate of 0.5, and label smoothing of 0.1. As summarized in~\cref{tab:hyperparameters}, ResNet-50~\cite{he2016deep} is trained with Adam for 5,000 steps, whereas DeiT-S/B~\cite{touvron2021training} and VMamba-T/S/B~\cite{liu2024vmamba} use AdamW with a weight decay of 0.05 for 10,000 steps.

To preserve pre-trained backbone representations while facilitating adaptation of the proposed modules, we use separate learning rates of $5\times10^{-5}$ for the backbone and $1\times10^{-4}$ for CPCNet and PP-CPCNet. For both methods, we set $\alpha_{\text{task}}=1$, with alignment weights $\alpha_{\text{cpca}}=\alpha_{\text{ppcpc}}=10^{-2}$. For subsequent experiments, we use the optimal configuration of $d=256$ and $T=3$ for CPCNet, and $T=1$ and $d_1=128$ for PP-CPCNet. The effects of these architectural choices are discussed in~\cref{ssec:ablation_analysis}. All experiments are conducted on NVIDIA A100 GPUs with 80\,GB of memory. As summarized in~\cref{tab:overall_summary_efficiency}, CPCNet and PP-CPCNet achieve competitive performance against SOTA methods across different backbones, while maintaining computational efficiency comparable to the ERM baseline. Domain-wise results across all benchmarks are additionally provided in the supplementary material (\cref{tab:pacs_vlcs,tab:officehome_terraincognita}) for a more detailed comparison with prior work.

\begin{table}[t]
\centering
\caption{Hyperparameter settings used across all experiments.}
\label{tab:hyperparameters}
\resizebox{\linewidth}{!}{%
\begin{tabular}{l c c c c c}
\toprule
\textbf{Backbone} & 
\makecell{\textbf{Feature} \\ \textbf{Dim.} (\(D\))} & 
\textbf{Optimizer} & 
\makecell{\textbf{Weight} \\ \textbf{Decay}} & 
\makecell{\textbf{Batch Size} \\ \textbf{per Domain}} & 
\makecell{\textbf{Training} \\ \textbf{Steps}} \\
\midrule
ResNet-50 & 2048 & Adam & 0 & \multirow{6}{*}{32} & 5,000 \\
\cmidrule(lr){1-4} \cmidrule(lr){6-6}
DeiT-S    & 384  & \multirow{5}{*}{AdamW} & \multirow{5}{*}{0.05} & & \multirow{5}{*}{10,000} \\
DeiT-B    & 768  & & & & \\
VMamba-T  & 768  & & & & \\
VMamba-S  & 768  & & & & \\
VMamba-B  & 1024 & & & & \\
\bottomrule
\end{tabular}%
}
\end{table}

\begin{table*}[t]
\centering
\caption{Summary of DG Accuracy (\%) and Computational Overhead Across Benchmarks. Total GPU time denotes the cumulative training time across three random seeds and four datasets. Best accuracy for each backbone is highlighted in \textbf{bold}. All results are reproduced under a unified training protocol, with gray rows indicating the proposed methods.}
\label{tab:overall_summary_efficiency}
\resizebox{\linewidth}{!}{%
\setlength{\tabcolsep}{8pt}
\begin{tabular}{p{4cm} | c c c c c | c c}
\toprule
\textbf{Method} & \texttt{PACS} & \texttt{VLCS} & \texttt{OfficeHome} & \texttt{TerraIncognita} & \textbf{Avg.} ($\uparrow$) & \makecell{\textbf{Peak GPU} \\ \textbf{RAM} (GB)} ($\downarrow$) & \makecell{\textbf{Total GPU Time} \\ (days-hh:mm:ss)} ($\downarrow$) \\
\midrule
\multicolumn{8}{c}{\hspace{4.3cm}\textbf{ResNet-50 Backbone}} \\
\midrule
ERM~\cite{vapnik1998statistical} \hfill \tiny{(Book '98)}  & 84.1 $\pm$ 0.3 & 74.5 $\pm$ 0.5 & 67.1 $\pm$ 0.3 & 49.0 $\pm$ 0.8 & 68.7 & \:\:8.15 & 0-16:18:01 \\
CORAL~\cite{sun2016deep} \hfill \tiny{(ECCV '16)}      & 82.9 $\pm$ 1.1 & 75.6 $\pm$ 0.4 & 67.3 $\pm$ 0.2 & 41.6 $\pm$ 1.1 & 66.9 & \:\:8.24 & 0-17:01:36 \\
CDANN~\cite{li2018deep} \hfill \tiny{(ECCV '18)}      & 83.6 $\pm$ 0.3 & 76.2 $\pm$ 0.4 & 66.7 $\pm$ 0.3 & 48.9 $\pm$ 0.7 & 68.9 & \:\:8.22 & 0-16:28:10  \\
SelfReg~\cite{kim2021selfreg} \hfill \tiny{(ICCV '21)} & 82.6 $\pm$ 0.7 & 76.2 $\pm$ 0.5 & 67.2 $\pm$ 0.2 & 47.2 $\pm$ 1.0 & 68.3 & \:\:8.24 & 0-17:14:25 \\
SagNet~\cite{nam2021reducing} \hfill \tiny{(CVPR '21)} & 82.7 $\pm$ 1.0 & 73.8 $\pm$ 0.6 & 64.8 $\pm$ 0.4 & 47.7 $\pm$ 0.5 & 67.2 & \:\:8.21 & 1-02:44:50 \\
ARM~\cite{zhang2021adaptive} \hfill \tiny{(NeurIPS '21)}  & 84.1 $\pm$ 0.4 & 75.8 $\pm$ 0.3 & 67.0 $\pm$ 0.3 & 45.4 $\pm$ 0.6 & 68.1 & 12.85 & 0-22:21:43 \\
IB-ERM~\cite{ahuja2021invariance} \hfill \tiny{(NeurIPS '21)}   & 82.9 $\pm$ 0.5 & 74.3 $\pm$ 0.6 & 65.6 $\pm$ 0.3 & \best{50.3}{0.7} & 68.3 & \:\:8.15 & \textbf{0-16:15:01} \\
IB-IRM~\cite{ahuja2021invariance} \hfill \tiny{(NeurIPS '21)}   & 82.0 $\pm$ 0.8 & 76.4 $\pm$ 0.6 & 58.1 $\pm$ 0.2 & 42.8 $\pm$ 1.1 & 64.8 & \:\:8.15 & 0-16:26:45 \\
Transfer~\cite{zhang2021quantifying} \hfill \tiny{(NeurIPS '21)}   & 83.0 $\pm$ 0.4 & 73.6 $\pm$ 0.7 & 62.5 $\pm$ 0.3 & 36.3 $\pm$ 1.3 & 63.9 & \:\:8.15 & 3-03:42:55 \\
EQRM~\cite{eastwood2022probable} \hfill \tiny{(NeurIPS '22)}  & 84.0 $\pm$ 0.7 & 76.0 $\pm$ 0.4 & 67.4 $\pm$ 0.2 & 45.0 $\pm$ 0.8 & 68.1 & \:\:\textbf{8.12} & 0-17:07:21 \\
ADRMX~\cite{demirel2023adrmx} \hfill \tiny{(arXiv '23)}  & 84.6 $\pm$ 0.3 & 75.6 $\pm$ 0.5 & 66.6 $\pm$ 0.2 & 47.2 $\pm$ 0.8 & 68.5 & 16.36 & 0-23:49:41 \\
RDM~\cite{nguyen2024domain} \hfill \tiny{(WACV '24)}  & 85.3 $\pm$ 0.3 & 74.6 $\pm$ 0.4 & 68.0 $\pm$ 0.2 & 49.5 $\pm$ 0.6 & 69.4 & \:\:8.15 & 0-16:19:27 \\
URM~\cite{krishnamachari2024uniformly} \hfill \tiny{(TMLR '24)}  & 81.4 $\pm$ 0.7 & \best{77.1}{0.2} & 65.3 $\pm$ 0.3 & 49.8 $\pm$ 0.9 & 68.4 & \:\:8.19 & 0-16:18:08 \\
\rowcolor{gray!10}
CPCNet                                             & \best{85.5}{0.3} & 75.9 $\pm$ 0.5 & \best{69.3}{0.3} & 47.4 $\pm$ 0.9 & \textbf{69.5} & \:\:8.65 & 0-16:49:40 \\
\rowcolor{gray!20}
PP-CPCNet                                             & 85.1 $\pm$ 0.3 & 75.3 $\pm$ 0.6 & 69.1 $\pm$ 0.2 & 47.5 $\pm$ 1.0 & 69.2 & \:\:8.25 & 0-16:47:22 \\
\midrule
\multicolumn{8}{c}{\hspace{4.3cm}\textbf{ViT-based Backbones}} \\
\midrule
\rowcolor{gray!10}
CPCNet-S                                             & 85.4 $\pm$ 0.4 & 78.4 $\pm$ 0.4 & 72.8 $\pm$ 0.4 & \best{44.7}{0.8} & 70.3 & \:\:7.17 & 1-20:53:26 \\
\rowcolor{gray!20}
PP-CPCNet-S                                             & 87.4 $\pm$ 0.4 & 78.0 $\pm$ 0.3 & 73.1 $\pm$ 0.4 & 42.4 $\pm$ 0.7 & 70.2 & \:\:\textbf{6.79} & \textbf{1-19:38:36} \\
\rowcolor{gray!10}
CPCNet-B                                             & \best{89.6}{0.3} & \best{79.8}{0.4} & 77.1 $\pm$ 0.2 & 44.6 $\pm$ 0.7 & \textbf{72.8} & 14.28 & 4-18:12:24 \\
\rowcolor{gray!20}
PP-CPCNet-B                                             & 89.5 $\pm$ 0.4 & 79.5 $\pm$ 0.5 & \best{77.6}{0.2} & 43.8 $\pm$ 0.8 & 72.6 & 13.90 & 4-17:03:21 \\
\midrule
\multicolumn{8}{c}{\hspace{4.3cm}\textbf{SSM-based Backbones}} \\
\midrule
\rowcolor{gray!10}
CPCNet-T                                           & 89.6 $\pm$ 0.7 & 79.1 $\pm$ 0.5 & 70.2 $\pm$ 0.4 & 54.1 $\pm$ 0.4 & 73.3 & 21.43 & 2-02:34:26 \\
\rowcolor{gray!20}
PP-CPCNet-T                                           & 89.8 $\pm$ 0.4 & 79.4 $\pm$ 0.4 & 69.1 $\pm$ 0.5 & 53.9 $\pm$ 1.5 & 73.1 & \textbf{18.41} & \textbf{2-01:08:39} \\
\rowcolor{gray!10}
CPCNet-S                                           & 92.6 $\pm$ 0.4 & 80.1 $\pm$ 0.5 & 75.4 $\pm$ 0.2 & 56.4 $\pm$ 0.3 & 76.1 & 37.41 & 3-18:05:48 \\
\rowcolor{gray!20}
PP-CPCNet-S                                           & 92.5 $\pm$ 0.6 & 79.9 $\pm$ 0.3 & 75.6 $\pm$ 0.5 & 56.8 $\pm$ 0.5 & 76.2 & 37.22 & 3-16:20:05 \\
\rowcolor{gray!10}
CPCNet-B                                           & 91.2 $\pm$ 0.6 & 79.6 $\pm$ 0.4 & \best{78.1}{0.3} & \best{57.7}{0.8} & 76.6 & 47.42 & 4-12:57:31 \\
\rowcolor{gray!20}
PP-CPCNet-B                                           & \best{92.9}{0.2} & \best{80.3}{0.4} & 77.8 $\pm$ 0.3 & \best{57.7}{0.9} & \textbf{77.2} & 47.81 & 4-11:23:51 \\
\bottomrule
\end{tabular}%
}
\end{table*}

\subsection{Ablation Studies and Analysis}
\label{ssec:ablation_analysis}

First, we examine the sensitivity of CPCNet and PP-CPCNet to their architectural hyperparameters using a ResNet-50 backbone. For CPCNet, we perform a grid search over the projection dimension $d$ and the number of unfolding stages $T$, selecting $d=256$ and $T=3$. For PP-CPCNet, we first vary $d_1$ with $T=1$ fixed, then search over $T$ with the selected $d_1$ fixed, yielding $d_1=128$ and $T=1$. These configurations are used for all subsequent experiments across backbones and datasets. Detailed results are provided in the supplementary material (\cref{tab:cpcnet_ablation,tab:ablation_d1,tab:ppcpcnet_ablation}).
We further assess the effect of feature modulation through the comparison between ERM and CPCNet. Since the final layers of both modulation MLPs are zero-initialized, the modulation initially reduces to an identity mapping. Disabling feature modulation therefore recovers the ERM baseline, allowing the performance difference to reflect the contribution of the proposed CPCNet components.

Besides, we omit empirical results for the naive CPCA-based approach introduced in~\cref{ssec:problem_formulation}, as repeatedly executing the conventional FG iterative solver during network training incurs prohibitive computational overhead. Lastly, CPCNet and PP-CPCNet achieve competitive DG performance; their improvements over existing methods are relatively modest in some settings. Our objective, however, is not to introduce another backbone architecture, but to explore a fundamental, complementary perspective on DG. By incorporating the statistical concept of CPC into deep learning, we demonstrate an alignment strategy that is broadly applicable across diverse architectures.
\section{Conclusion}
\label{sec:conclusion}

In this paper, we presented a principled approach to DG by integrating CPCA into deep learning. To enable end-to-end differentiable optimization, we developed two complementary frameworks. CPCNet unfolds the FG algorithm into differentiable neural layers, while PP-CPCNet adopts a PP-based formulation that directly optimizes a common orthogonal basis, mitigating rank deficiency in small-sample regimes.
Experiments on joint diagonalization show that both frameworks closely recover the solutions of classical CPCA, while evaluations across four DomainBed benchmarks demonstrate their effectiveness for zero-shot DG across diverse backbone architectures. Overall, this work establishes a principled connection between classical multivariate statistics and end-to-end deep learning, highlighting the potential of CPCA as a complementary approach to learning transferable representations for DG.
\section*{LLM Usage Statement}
\addcontentsline{toc}{section}{LLM Usage Statement}

The authors used OpenAI's ChatGPT during manuscript preparation to improve language quality, refine textual presentation, and assist with the organization and refinement of related work and citations. The tool also helped refine code and clarify technical implementation details. These uses were limited to assisting with existing author-developed material and did not involve formulating research ideas or hypotheses, developing the methodology, designing or conducting experiments, analyzing experimental results, or drawing scientific conclusions. The authors did not use generative AI tools to prepare figures or visual materials, or to generate data, create datasets, or annotate. All AI-assisted suggestions and outputs were carefully reviewed and independently verified by the authors, who retain full responsibility for the submitted manuscript and its scientific content.
{
    \addcontentsline{toc}{section}{References}
    \small
    \bibliographystyle{ieeenat_fullname}    
    \bibliography{main}
}

\clearpage
\setcounter{page}{1}
\maketitlesupplementary

\begin{strip}
\tableofcontents
\addcontentsline{toc}{section}{Supplementary Material}
\end{strip}
\clearpage

\begin{figure*}[!t]
    \centering
    \includegraphics[width=\linewidth]{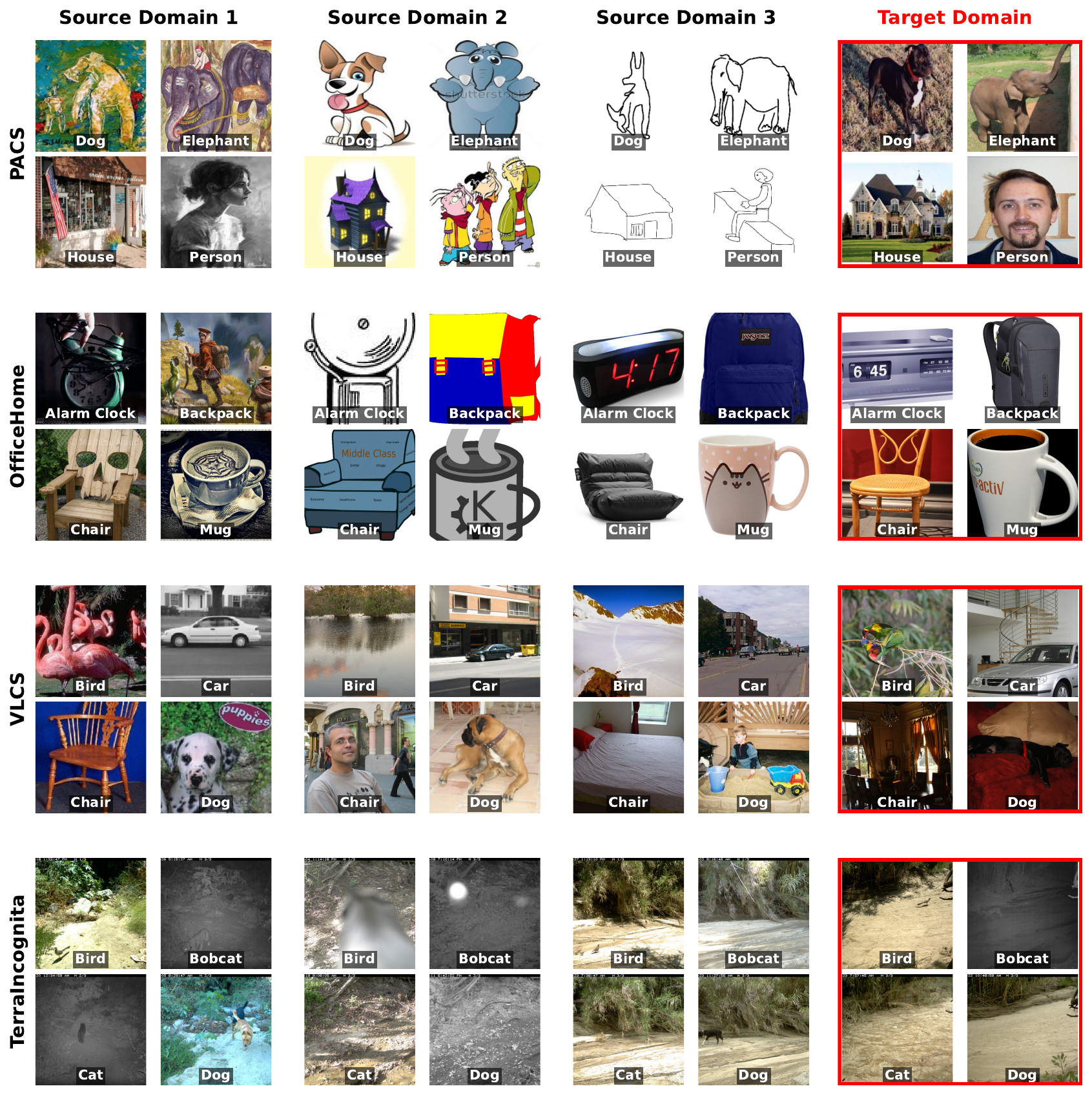}
    \caption{Visual examples from the four DG benchmarks: \texttt{PACS}, \texttt{VLCS}, \texttt{OfficeHome}, and \texttt{TerraIncognita}.}
    \label{fig:DG_benchmarks}
\end{figure*}

\begin{algorithm*}[t]
\caption{Forward Pass and Objective Formulation of CPCNet}
\label{alg:cpcnet}
\textbf{Input:} Mini-batch of raw inputs $\mathbf{X}$, domain labels $\mathbf{d} \in \{1, \dots, K\}$, ground-truth labels $\mathbf{Y}$ \\
\textbf{Parameters:} Backbone $h_\theta$, Bottleneck $b_\psi$, Hypernetwork $H_\phi$, $\mathrm{MLP}_1$, $\mathrm{MLP}_2$, Classifier $(\mathbf{W}_{\text{cls}}, \mathbf{b}_{\text{cls}})$ \\
\textbf{Output:} Task prediction logits $\hat{\mathbf{Y}}$, CPC regularization loss $\mathcal{L}_{\text{cpc}}$, and total objective $\mathcal{L}_{\text{total}}$

\vspace{2mm}
\tcc{1. Feature Extraction \& Covariance Estimation}
$\mathbf{F} \leftarrow h_\theta(\mathbf{X})$ \tcp*{Extract $D$-dim backbone features}
$\mathbf{Z} \leftarrow b_\psi(\mathbf{F})$ \tcp*{Project to $d$-dim bottleneck representation}
\For{$k = 1, \dots, K$}{
    $\mathbf{S}_k \leftarrow \mathrm{Cov}(\mathbf{Z}[\mathbf{d} = k])$ \tcp*{Compute latent domain covariances}
}

\vspace{2mm}
\tcc{2. Domain-Adaptive Unfolding (\cref{sssec:dynamic_unfolding})}
$\boldsymbol{\eta} \leftarrow \frac{1}{2} \cdot \sigma\Big(H_\phi(\mathbf{S}_1, \dots, \mathbf{S}_K)\Big)$ \tcp*{Predict context-dependent step-sizes $\eta_1, \dots, \eta_T$}

\vspace{2mm}
\tcc{3. Deep Unfolded CPCA Solver (\cref{sssec:orthogonal_cayley,sssec:riemannian_gradient})}
$\mathbf{A}_0 \leftarrow \mathbf{0}_{d \times d}$ \tcp*{Initialize skew-symmetric tangent space matrix}
\For{$t = 1, \dots, T$}{
    $\boldsymbol{\beta}_{t-1} \leftarrow \left(\mathbf{I}_d - \frac{1}{2}\mathbf{A}_{t-1}\right)\left(\mathbf{I}_d + \frac{1}{2}\mathbf{A}_{t-1}\right)^{-1}$ \tcp*{Map to orthogonal group via Cayley transform}
    
    \For{$k = 1, \dots, K$}{
        $\hat{\mathbf{S}}_k \leftarrow \boldsymbol{\beta}_{t-1}^\top \mathbf{S}_k \boldsymbol{\beta}_{t-1}$ \tcp*{Transform covariances to the current basis}
        $\boldsymbol{\Omega}_k[l,m] \leftarrow \frac{(\hat{\mathbf{S}}_k)_{ll} - (\hat{\mathbf{S}}_k)_{mm}}{(\hat{\mathbf{S}}_k)_{ll}(\hat{\mathbf{S}}_k)_{mm} + \epsilon}$ \tcp*{Compute skew-symmetric weight matrix}
    }
    
    $\mathbf{G}_{\mathbf{A}} \leftarrow \sum_{k=1}^K n_k \big( \hat{\mathbf{S}}_k \odot \boldsymbol{\Omega}_k \big)$ \tcp*{Compute Riemannian gradient in tangent space}
    $\tilde{\mathbf{G}}_{\mathbf{A}} \leftarrow \mathbf{G}_{\mathbf{A}} \,/\, (\|\mathbf{G}_{\mathbf{A}}\|_F + \epsilon')$ \tcp*{Normalize gradient to stabilize unfolding}
    
    $\mathbf{A}_t \leftarrow \mathbf{A}_{t-1} - \eta_t \tilde{\mathbf{G}}_{\mathbf{A}}$ \tcp*{Update skew-symmetric matrix}
}

\vspace{2mm}
\tcc{4. Common Orthogonal Basis \& Feature Modulation (\cref{sssec:mgfm})}
$\boldsymbol{\beta}_T \leftarrow \left(\mathbf{I}_d - \frac{1}{2}\mathbf{A}_T\right)\left(\mathbf{I}_d + \frac{1}{2}\mathbf{A}_T\right)^{-1}$ \tcp*{Extract the final common orthogonal basis}

$\mathbf{U} \leftarrow \mathbf{Z} \boldsymbol{\beta}_T$ \tcp*{Project bottleneck features onto the common basis}

$\boldsymbol{\gamma} \leftarrow 2 \cdot \sigma\Big(\mathrm{MLP}_1(\mathbf{U})\Big)$ \tcp*{Generate domain-invariant scaling factors}
$\Delta \mathbf{F} \leftarrow \mathrm{MLP}_2(\mathbf{U})$ \tcp*{Generate domain-invariant shifting factors}

$\tilde{\mathbf{F}} \leftarrow (\mathbf{F} \odot \boldsymbol{\gamma}) + \Delta \mathbf{F}$ \tcp*{Modulate the high-dimensional ambient features}

\vspace{2mm}
\tcc{5. Objective Computation (\cref{sssec:cpcnet_objective})}
$\hat{\mathbf{Y}} \leftarrow \tilde{\mathbf{F}} \mathbf{W}_{\text{cls}} + \mathbf{b}_{\text{cls}}$ \tcp*{Compute task prediction logits for classification}

\For{$k = 1, \dots, K$}{
    $\hat{\mathbf{S}}_k^{(T)} \leftarrow \boldsymbol{\beta}_T^\top \mathbf{S}_k \boldsymbol{\beta}_T$ \tcp*{Recompute transformed covariances with final basis}
}

$\mathcal{L}_{\text{cpc}} \leftarrow \frac{1}{K} \sum_{k=1}^K \frac{\|\hat{\mathbf{S}}_k^{(T)}\|_F^2 - \|\mathrm{diag}(\hat{\mathbf{S}}_k^{(T)})\|_F^2}{d(d-1)}$ \tcp*{Compute CPC off-diagonal penalty}

$\mathcal{L}_{\text{total}} \leftarrow \alpha_{\text{task}}\mathcal{L}_{\text{task}}(\hat{\mathbf{Y}}, \mathbf{Y}) + \alpha_{\text{cpc}} \mathcal{L}_{\text{cpc}}$ \tcp*{Compute final weighted training objective}

\textbf{return} $\hat{\mathbf{Y}}, \mathcal{L}_{\text{cpc}}, \mathcal{L}_{\text{total}}$
\end{algorithm*}

\clearpage


\begin{algorithm*}[t]
\caption{Forward Pass and Objective Formulation of PP-CPCNet}
\label{alg:pp_cpcnet}
\textbf{Input:} Mini-batch of raw inputs $\mathbf{X}$, domain labels $\mathbf{d} \in \{1, \dots, K\}$ with sizes $N_k$, ground-truth labels $\mathbf{Y}$ \\
\textbf{Parameters:} Backbone $h_\theta$, Cascade depth $T$, Dimensionalities $\{d_t\}_{t=1}^T$, Depth-specific Bottlenecks $b_{\psi_t}$, \\
\hspace*{1.87cm} Global Manifold parameters $\{\boldsymbol{\theta}_t\}_{t=1}^T$, Hypernetworks $\mathrm{MLP}_1$, $\mathrm{MLP}_2$, Classifier $(\mathbf{W}_{\text{cls}}, \mathbf{b}_{\text{cls}})$ \\
\textbf{Output:} Task prediction logits $\hat{\mathbf{Y}}$, PP regularization loss $\mathcal{L}_{\text{ppcpc}}$, Optimization objective $\mathcal{L}_{\text{total}}$

\vspace{2mm}
\tcc{1. Initial Feature Extraction}
$\tilde{\mathbf{F}}_0 \leftarrow h_\theta(\mathbf{X})$ \tcp*{Extract ambient backbone features}

\vspace{2mm}
\tcc{2. Optimization \& Modulation (\cref{sssec:parametric_stiefel,sssec:mgfm})}
\For{$t = 1, \dots, T$}{
    \vspace{0.5mm}
    $\mathbf{Z}_t \leftarrow b_{\psi_t}(\tilde{\mathbf{F}}_{t-1})$ \tcp*{Project to narrowed $d_t$-dim bottleneck}
    $\mathbf{A}_t \leftarrow \mathcal{S}(\boldsymbol{\theta}_t)$ \tcp*{Construct global skew-symmetric matrix}
    $\boldsymbol{\beta}_t \leftarrow \left(\mathbf{I}_{d_t} - \frac{1}{2}\mathbf{A}_t\right)\left(\mathbf{I}_{d_t} + \frac{1}{2}\mathbf{A}_t\right)^{-1}$ \tcp*{Map to Stiefel Manifold via Cayley Transform}
    $\mathbf{U}_t \leftarrow \mathbf{Z}_t \boldsymbol{\beta}_t$ \tcp*{Project features onto the orthogonal basis}
    $\boldsymbol{\gamma}_t \leftarrow 2 \cdot \sigma\Big(\mathrm{MLP}_1(\mathbf{U}_t)\Big)$ \tcp*{Generate domain-invariant scaling}
    $\Delta \mathbf{F}_t \leftarrow \mathrm{MLP}_2(\mathbf{U}_t)$ \tcp*{Generate domain-invariant shift}
    $\tilde{\mathbf{F}}_t \leftarrow (\tilde{\mathbf{F}}_{t-1} \odot \boldsymbol{\gamma}_t) + \Delta \mathbf{F}_t$ \tcp*{Modulate ambient features residually}
}

\vspace{2mm}
\tcc{3. Symmetry-Breaking Robust Dispersion (\cref{sssec:symmetry_breaking})}
$\mathcal{L}_{\text{ppcpc}} \leftarrow 0$ \\
\For{$t = 1, \dots, T$}{
    \For{$k = 1, \dots, K$}{
        $\mathbf{U}_{t,k} \leftarrow \mathbf{U}_t[\mathbf{d} = k]$ \tcp*{Isolate depth-$t$ projections for domain $k$}
        
        $\mathbf{m}_{t,k} \leftarrow \mathrm{sg}\big[ \mathrm{median}\big( \{ \mathbf{U}_{t,k}^{(i)} \}_{i=1}^{N_k} \big) \big]$ \tcp*{Fix robust median anchor via stop-gradient}
        
        $s_{t,k} \leftarrow \mathbf{w}_t^\top \left( \frac{1}{N_k} \sum_{i=1}^{N_k} \big| \mathbf{U}_{t,k}^{(i)} - \mathbf{m}_{t,k} \big| \right)$ \tcp*{Compute dense and weighted L1 dispersion}
        
        $\mathcal{L}_{\text{ppcpc}} \leftarrow \mathcal{L}_{\text{ppcpc}} - s_{t,k}$ \tcp*{Accumulate negative score to maximize variance}
    }
}
$\mathcal{L}_{\text{ppcpc}} \leftarrow \mathcal{L}_{\text{ppcpc}} \,/\, (K \cdot T)$ \tcp*{Normalize objective across domains and depths}

\vspace{2mm}
\tcc{4. Objective Formulation and Inference (\cref{sssec:ofi})}
$\hat{\mathbf{Y}} \leftarrow \tilde{\mathbf{F}}_T \mathbf{W}_{\text{cls}} + \mathbf{b}_{\text{cls}}$ \tcp*{Compute task prediction logits from final depth}
$\mathcal{L}_{\text{total}} \leftarrow \alpha_{\text{task}}\mathcal{L}_{\text{task}}(\hat{\mathbf{Y}}, \mathbf{Y}) + \alpha_{\text{ppcpc}} \mathcal{L}_{\text{ppcpc}}$ \tcp*{Formulate final joint objective}

\textbf{return} $\hat{\mathbf{Y}}, \mathcal{L}_{\text{ppcpc}}, \mathcal{L}_{\text{total}}$
\end{algorithm*}

\clearpage

\begin{table*}[t]
\centering
\caption{Detailed statistics of the DG benchmark datasets.}
\label{tab:detailed_dataset_summary}
\resizebox{\linewidth}{!}{%
\setlength{\tabcolsep}{8pt}
\begin{tabular}{l l c c c @{\hspace{1.5em}} r@{\hspace{0.3em}}l@{\hspace{0.8em}}c@{\hspace{0.8em}}r@{\hspace{0.3em}}l @{\hspace{1.5em}} r@{\hspace{0.15em}}c@{\hspace{0.15em}}l@{\hspace{0.7em}}c@{\hspace{0.7em}}r@{\hspace{0.15em}}c@{\hspace{0.15em}}l}
\toprule
\textbf{Dataset} & \textbf{Domain} & \makecell{\textbf{Number} \\ \textbf{of Images}} & \makecell{\textbf{Total Number} \\ \textbf{of Images}} & \makecell{\textbf{Number} \\ \textbf{of Classes}} & \multicolumn{5}{c}{\textbf{Range of Images per Class}} & \multicolumn{7}{c}{\textbf{Range of Resolutions}} \\
\midrule
\multirow{4}{*}{\texttt{PACS}~\cite{li2017deeper}$^1$} 
              & Art         & 2,048 & \multirow{4}{*}{9,991}  & \multirow{4}{*}{7} & 184 & (guitar)  & to & 449 & (person)   & \multicolumn{7}{c}{\multirow{4}{*}{$227{\times}227$}} \\
              & Cartoon     & 2,344 &                         &                    & 135 & (guitar)  & to & 457 & (elephant) & \multicolumn{7}{c}{}                                         \\
              & Photo       & 1,670 &                         &                    & 182 & (giraffe) & to & 432 & (person)   & \multicolumn{7}{c}{}                                         \\
              & Sketch      & 3,929 &                         &                    & 80  & (house)   & to & 816 & (horse)    & \multicolumn{7}{c}{}                                         \\
\midrule
\multirow{4}{*}{\texttt{VLCS}~\cite{fang2013unbiased}$^2$} 
              & Caltech101  & 1,415 & \multirow{4}{*}{10,729} & \multirow{4}{*}{5} & 67  & (dog)     & to & 870  & (person) & 80  & $\times$ & 174  & to & 708  & $\times$ & 468  \\
              & LabelMe     & 2,656 &                         &                    & 42  & (dog)     & to & 1237 & (person) & 375 & $\times$ & 384  & to & 3504 & $\times$ & 4257 \\
              & SUN09       & 3,282 &                         &                    & 20  & (bird)    & to & 1264 & (person) & 187 & $\times$ & 174  & to & 6144 & $\times$ & 4096 \\
              & VOC2007     & 3,376 &                         &                    & 330 & (bird)    & to & 1499 & (person) & 200 & $\times$ & 97   & to & 500  & $\times$ & 500  \\
\midrule
\multirow{4}{*}{\texttt{OfficeHome}~\cite{venkateswara2017deep}$^3$} 
              & Art         & 2,427 & \multirow{4}{*}{15,588} & \multirow{4}{*}{65}& 15  & (drill)         & to & 99 & (bottle)   & 85  & $\times$ & 117  & to & 4438 & $\times$ & 2686 \\
              & Clipart     & 4,365 &                         &                    & 39  & (toothbrush)    & to & 99 & (scissors) & 18  & $\times$ & 4    & to & 2400 & $\times$ & 2400 \\
              & product     & 4,439 &                         &                    & 38  & (postit notes) & to & 99 & (scissors) & 42  & $\times$ & 42   & to & 2560 & $\times$ & 2560 \\
              & Real World  & 4,357 &                         &                    & 23  & (marker)        & to & 99 & (candles)  & 80  & $\times$ & 63   & to & 6000 & $\times$ & 6500 \\
\midrule
\multirow{4}{*}{\texttt{TerraIncognita}~\cite{beery2018recognition}$^4$} 
              & L100        & 4,879 & \multirow{4}{*}{24,788} & \multirow{4}{*}{10}& 12 & (coyote)  & to & 2552 & (raccoon) & \multicolumn{7}{c}{\multirow{4}{*}{$1024{\times}747$}} \\
              & L38         & 9,767 &                         &                    & 3  & (bird)    & to & 4500 & (opossum) & \multicolumn{7}{c}{}                                          \\
              & L43         & 4,020 &                         &                    & 3  & (cat)     & to & 1104 & (bobcat)  & \multicolumn{7}{c}{}                                          \\
              & L46         & 6,122 &                         &                    & 14 & (squirrel)& to & 1464 & (raccoon) & \multicolumn{7}{c}{}                                          \\
\bottomrule
\end{tabular}%
}
\vspace{1mm}
\raggedright
\scriptsize
$^1$\texttt{PACS} dataset acquired from: \url{https://huggingface.co/datasets/flwrlabs/pacs} \\
$^2$\texttt{VLCS} dataset acquired from: \url{https://www.kaggle.com/datasets/iamjanvijay/vlcsdataset} \\
$^3$\texttt{OfficeHome} dataset acquired from: \url{https://www.hemanthdv.org/officeHomeDataset.html} \\
$^4$\texttt{TerraIncognita} dataset acquired from: \url{https://storage.googleapis.com/public-datasets-lila/caltechcameratraps/eccv_18_all_images_sm.tar.gz}
\end{table*}

\clearpage


\begin{table*}[t]
\centering
\caption{Detailed DG accuracies (\%) on \texttt{PACS} and \texttt{VLCS}. Best results within each backbone category are highlighted in \textbf{bold}. N/A denotes unreported results or standard deviations. Methods marked with $^\dagger$ are from the unified benchmark framework~\cite{gulrajanisearch}; other results are collected from the original papers or supplementary materials. Gray rows indicate results reproduced by us.}
\label{tab:pacs_vlcs}
\resizebox{\linewidth}{!}{%
\begin{tabular}{p{4cm} | C C C C | c || C C C C | c }
\toprule
\multirow{2.5}{*}{\textbf{Method}} & \multicolumn{5}{c||}{\texttt{PACS}} & \multicolumn{5}{c}{\texttt{VLCS}} \\
\cmidrule(lr){2-6} \cmidrule(lr){7-11}
 & Art & Cartoon & Photo & Sketch & Avg.($\uparrow$) & Caltech & LabelMe & Pascal & Sun & Avg. ($\uparrow$) \\
\midrule
\multicolumn{11}{c}{\textbf{ResNet-50 Backbone}} \\
\midrule
ERM$^\dagger$~\cite{vapnik1998statistical} \hfill \tiny{(Book '98)} & 84.7 $\pm$ 0.4 & 80.8 $\pm$ 0.6 & 97.2 $\pm$ 0.3 & 79.3 $\pm$ 1.0 & 85.5 $\pm$ 0.2 & 97.7 $\pm$ 0.4 & 64.3 $\pm$ 0.9 & 73.4 $\pm$ 0.5 & 74.6 $\pm$ 1.3 & 77.5 $\pm$ 0.4 \\
\rowcolor{gray!10}
ERM (reproduced) & 81.8 $\pm$ 0.8 & 82.6 $\pm$ 0.9 & 94.7 $\pm$ 0.3 & 77.4 $\pm$ 0.2 & 84.1 $\pm$ 0.3 & 94.8 $\pm$ 0.6 & 62.6 $\pm$ 1.2 & 66.3 $\pm$ 1.2 & 74.5 $\pm$ 0.8 & 74.5 $\pm$ 0.5 \\
CORAL$^\dagger$~\cite{sun2016deep} \hfill \tiny{(ECCV '16)} & 88.3 $\pm$ 0.2 & 80.0 $\pm$ 0.5 & 97.5 $\pm$ 0.3 & 78.8 $\pm$ 1.3 & 86.2 $\pm$ 0.3 & 98.3 $\pm$ 0.1 & \best{66.1}{1.2} & 73.4 $\pm$ 0.3 & 77.5 $\pm$ 1.2 & 78.8 $\pm$ 0.6 \\
\rowcolor{gray!10}
CORAL (reproduced) & 80.0 $\pm$ 1.4 & 82.7 $\pm$ 1.1 & 93.4 $\pm$ 0.7 & 75.7 $\pm$ 3.9 & 82.9 $\pm$ 1.1 & 96.9 $\pm$ 0.6 & 60.3 $\pm$ 1.0 & 71.3 $\pm$ 0.8 & 73.8 $\pm$ 1.0 & 75.6 $\pm$ 0.4 \\
CDANN~\cite{li2018deep} \hfill \tiny{(ECCV '18)} & \small{N/A}\: $\pm$ \footnotesize{N/A} & \small{N/A}\: $\pm$ \footnotesize{N/A} & \small{N/A}\: $\pm$ \footnotesize{N/A} & \small{N/A}\: $\pm$ \footnotesize{N/A} & \small{N/A}\: $\pm$ \footnotesize{N/A} & \small{N/A}\: $\pm$ \footnotesize{N/A} & \small{N/A}\: $\pm$ \footnotesize{N/A} & \small{N/A}\: $\pm$ \footnotesize{N/A} & \small{N/A}\: $\pm$ \footnotesize{N/A} & \small{N/A}\: $\pm$ \footnotesize{N/A} \\
\rowcolor{gray!10}
CDANN (reproduced) & 83.1 $\pm$ 1.0 & 79.8 $\pm$ 0.6 & 95.6 $\pm$ 0.4 & 76.1 $\pm$ 0.6 & 83.6 $\pm$ 0.3 & 96.1 $\pm$ 0.5 & 63.4 $\pm$ 0.4 & 70.3 $\pm$ 1.1 & 74.8 $\pm$ 0.9 & 76.2 $\pm$ 0.4 \\
DGER~\cite{zhao2020domain} \hfill \tiny{(NeurIPS '20)} & 87.5 $\pm$ 1.0 & 79.3 $\pm$ 1.4 & \best{98.3}{0.1} & 76.3 $\pm$ 0.7 & 85.3 $\pm$ \scriptsize{N/A} & \small{N/A}\: $\pm$ \footnotesize{N/A} & \small{N/A}\: $\pm$ \footnotesize{N/A} & \small{N/A}\: $\pm$ \footnotesize{N/A} & \small{N/A}\: $\pm$ \footnotesize{N/A} & \small{N/A}\: $\pm$ \footnotesize{N/A} \\
SelfReg~\cite{kim2021selfreg} \hfill \tiny{(ICCV '21)} & 87.9 $\pm$ 1.0 & 79.4 $\pm$ 1.4 & 96.8 $\pm$ 0.7 & 78.3 $\pm$ 1.2 & 85.6 $\pm$ 0.4 & 96.7 $\pm$ 0.4 & 65.2 $\pm$ 1.2 & 73.1 $\pm$ 1.3 & 76.2 $\pm$ 0.7 & 77.8 $\pm$ 0.9 \\
\rowcolor{gray!10}
SelfReg (reproduced) & 83.0 $\pm$ 0.8 & 77.8 $\pm$ 1.2 & 95.4 $\pm$ 0.8 & 74.2 $\pm$ 2.1 & 82.6 $\pm$ 0.7 & 96.5 $\pm$ 0.6 & 64.4 $\pm$ 1.0 & 69.4 $\pm$ 1.2 & 74.6 $\pm$ 1.1 & 76.2 $\pm$ 0.5 \\
SagNet$^\dagger$~\cite{nam2021reducing} \hfill \tiny{(CVPR '21)} & 87.4 $\pm$ 1.0 & 80.7 $\pm$ 0.6 & 97.1 $\pm$ 0.1 & 80.0 $\pm$ 0.4 & 86.3 $\pm$ 0.2 & 97.9 $\pm$ 0.4 & 64.5 $\pm$ 0.5 & 71.4 $\pm$ 1.3 & 77.5 $\pm$ 0.5 & 77.8 $\pm$ 0.5 \\
\rowcolor{gray!10}
SagNet (reproduced) & 80.8 $\pm$ 2.1 & 79.3 $\pm$ 2.1 & 95.0 $\pm$ 0.9 & 75.6 $\pm$ 2.7 & 82.7 $\pm$ 1.0 & 93.6 $\pm$ 2.0 & 61.6 $\pm$ 0.4 & 67.4 $\pm$ 0.6 & 72.6 $\pm$ 1.1 & 73.8 $\pm$ 0.6 \\
SWAD~\cite{cha2021swad} \hfill \tiny{(NeurIPS '21)} & 89.3 $\pm$ 0.2 & \best{83.4}{0.6} & 97.3 $\pm$ 0.3 & \best{82.5}{0.5} & 88.1 $\pm$ 0.1 & 98.8 $\pm$ 0.1 & 63.3 $\pm$ 0.3 & 75.3 $\pm$ 0.5 & 79.2 $\pm$ 0.6 & 79.1 $\pm$ 0.1 \\
ARM$^\dagger$~\cite{zhang2021adaptive} \hfill \tiny{(NeurIPS '21)} & 86.8 $\pm$ 0.6 & 76.8 $\pm$ 0.5 & 97.4 $\pm$ 0.3 & 79.3 $\pm$ 1.2 & 85.1 $\pm$ 0.4 & 98.7 $\pm$ 0.2 & 63.6 $\pm$ 0.7 & 71.3 $\pm$ 1.2 & 76.7 $\pm$ 0.6 & 77.6 $\pm$ 0.3 \\
\rowcolor{gray!10}
ARM (reproduced) & 83.9 $\pm$ 1.2 & 80.8 $\pm$ 0.3 & 94.8 $\pm$ 0.3 & 76.7 $\pm$ 1.2 & 84.1 $\pm$ 0.4 & 95.5 $\pm$ 0.6 & 64.8 $\pm$ 0.4 & 67.2 $\pm$ 0.5 & 75.6 $\pm$ 1.0 & 75.8 $\pm$ 0.3 \\
IB-ERM~\cite{ahuja2021invariance} \hfill \tiny{(NeurIPS '21)} & \small{N/A}\: $\pm$ \footnotesize{N/A} & \small{N/A}\: $\pm$ \footnotesize{N/A} & \small{N/A}\: $\pm$ \footnotesize{N/A} & \small{N/A}\: $\pm$ \footnotesize{N/A} & \small{N/A}\: $\pm$ \footnotesize{N/A} & \small{N/A}\: $\pm$ \footnotesize{N/A} & \small{N/A}\: $\pm$ \footnotesize{N/A} & \small{N/A}\: $\pm$ \footnotesize{N/A} & \small{N/A}\: $\pm$ \footnotesize{N/A} & \small{N/A}\: $\pm$ \footnotesize{N/A} \\
\rowcolor{gray!10}
IB-ERM (reproduced) & 82.7 $\pm$ 0.7 & 78.3 $\pm$ 1.8 & 94.3 $\pm$ 0.1 & 76.4 $\pm$ 1.0 & 82.9 $\pm$ 0.5 & 94.8 $\pm$ 1.0 & 62.5 $\pm$ 1.2 & 67.6 $\pm$ 1.5 & 72.4 $\pm$ 1.0 & 74.3 $\pm$ 0.6 \\
IB-IRM~\cite{ahuja2021invariance} \hfill \tiny{(NeurIPS '21)} & \small{N/A}\: $\pm$ \footnotesize{N/A} & \small{N/A}\: $\pm$ \footnotesize{N/A} & \small{N/A}\: $\pm$ \footnotesize{N/A} & \small{N/A}\: $\pm$ \footnotesize{N/A} & \small{N/A}\: $\pm$ \footnotesize{N/A} & \small{N/A}\: $\pm$ \footnotesize{N/A} & \small{N/A}\: $\pm$ \footnotesize{N/A} & \small{N/A}\: $\pm$ \footnotesize{N/A} & \small{N/A}\: $\pm$ \footnotesize{N/A} & \small{N/A}\: $\pm$ \footnotesize{N/A} \\
\rowcolor{gray!10}
IB-IRM (reproduced) & 81.9 $\pm$ 0.6 & 76.2 $\pm$ 0.3 & 95.9 $\pm$ 0.3 & 74.1 $\pm$ 2.9 & 82.0 $\pm$ 0.8 & 96.2 $\pm$ 0.3 & 63.3 $\pm$ 1.5 & 69.3 $\pm$ 1.0 & 76.9 $\pm$ 1.8 & 76.4 $\pm$ 0.6 \\
Transfer~\cite{zhang2021quantifying} \hfill \tiny{(NeurIPS '21)} & \small{N/A}\: $\pm$ \footnotesize{N/A} & \small{N/A}\: $\pm$ \footnotesize{N/A} & \small{N/A}\: $\pm$ \footnotesize{N/A} & \small{N/A}\: $\pm$ \footnotesize{N/A} & \small{N/A}\: $\pm$ \footnotesize{N/A} & \small{N/A}\: $\pm$ \footnotesize{N/A} & \small{N/A}\: $\pm$ \footnotesize{N/A} & \small{N/A}\: $\pm$ \footnotesize{N/A} & \small{N/A}\: $\pm$ \footnotesize{N/A} & \small{N/A}\: $\pm$ \footnotesize{N/A} \\
\rowcolor{gray!10}
Transfer (reproduced) & 79.5 $\pm$ 1.2 & 82.6 $\pm$ 0.8 & 92.6 $\pm$ 0.3 & 77.2 $\pm$ 0.7 & 83.0 $\pm$ 0.4 & 93.8 $\pm$ 1.8 & 62.5 $\pm$ 0.7 & 67.0 $\pm$ 0.5 & 71.2 $\pm$ 1.8 & 73.6 $\pm$ 0.7 \\
EQRM~\cite{eastwood2022probable} \hfill \tiny{(NeurIPS '22)} & 86.5 $\pm$ 0.4 & 82.1 $\pm$ 0.7 & 96.6 $\pm$ 0.2 & 80.8 $\pm$ 0.2 & 86.5 $\pm$ 0.2 & 98.3 $\pm$ 0.0 & 63.7 $\pm$ 0.8 & 72.6 $\pm$ 1.0 & 76.7 $\pm$ 1.1 & 77.8 $\pm$ 0.6 \\
\rowcolor{gray!10}
EQRM (reproduced) & 83.0 $\pm$ 0.6 & 80.8 $\pm$ 0.5 & 94.5 $\pm$ 0.5 & 77.7 $\pm$ 2.6 & 84.0 $\pm$ 0.7 & 97.1 $\pm$ 0.2 & 62.8 $\pm$ 1.4 & 70.2 $\pm$ 0.4 & 73.8 $\pm$ 1.0 & 76.0 $\pm$ 0.4 \\
EoA~\cite{arpit2022ensemble} \hfill \tiny{(NeurIPS '22)} & \best{90.5}{\scriptsize{N/A}} & \best{83.4}{\scriptsize{N/A}} & 98.0 $\pm$ \scriptsize{N/A} & \best{82.5}{\scriptsize{N/A}} & \best{88.6}{\scriptsize{N/A}} & \best{99.1}{\scriptsize{N/A}} & 63.1 $\pm$ \scriptsize{N/A} & \best{75.9}{\scriptsize{N/A}} & 78.3 $\pm$ \scriptsize{N/A} & 79.1 $\pm$ \scriptsize{N/A} \\
ADRMX~\cite{demirel2023adrmx} \hfill \tiny{(arXiv '23)} & 87.7 $\pm$ \scriptsize{N/A} & 80.6 $\pm$ \scriptsize{N/A} & 97.7 $\pm$ \scriptsize{N/A} & 77.5 $\pm$ \scriptsize{N/A} & 85.9 $\pm$ \scriptsize{N/A} & \small{N/A}\: $\pm$ \footnotesize{N/A} & \small{N/A}\: $\pm$ \footnotesize{N/A} & \small{N/A}\: $\pm$ \footnotesize{N/A} & \small{N/A}\: $\pm$ \footnotesize{N/A} & 78.5 $\pm$ \scriptsize{N/A} \\
\rowcolor{gray!10}
ADRMX (reproduced) & 86.7 $\pm$ 1.0 & 80.7 $\pm$ 0.5 & 96.2 $\pm$ 0.6 & 74.9 $\pm$ 0.3 & 84.6 $\pm$ 0.3 & 94.8 $\pm$ 1.5 & 62.7 $\pm$ 0.2 & 69.3 $\pm$ 0.9 & 75.6 $\pm$ 0.6 & 75.6 $\pm$ 0.5 \\
MADG~\cite{dayal2023madg} \hfill \tiny{(NeurIPS '23)} & 87.8 $\pm$ 0.5 & 82.2 $\pm$ 0.6 & 97.7 $\pm$ 0.3 & 78.3 $\pm$ 0.4 & 86.5 $\pm$ 0.4 & 98.5 $\pm$ 0.2 & 65.8 $\pm$ 0.3 & 73.1 $\pm$ 0.3 & 77.3 $\pm$ 0.1 & 78.7 $\pm$ 0.2 \\
SAGM~\cite{wang2023sharpness} \hfill \tiny{(CVPR '23)} & 87.4 $\pm$ 0.2 & 80.2 $\pm$ 0.3 & 98.0 $\pm$ 0.2 & 80.8 $\pm$ 0.6 & 86.6 $\pm$ 0.2 & 99.0 $\pm$ 0.2 & 65.2 $\pm$ 0.4 & 75.1 $\pm$ 0.3 & \best{80.7}{0.8} & \best{80.0}{0.3} \\
GMDG~\cite{tan2024rethinking} \hfill \tiny{(CVPR '24)} & 84.7 $\pm$ 1.0 & 81.7 $\pm$ 2.4 & 97.5 $\pm$ 0.4 & 80.5 $\pm$ 1.8 & 85.6 $\pm$ 0.3 & 98.3 $\pm$ 0.4 & 65.9 $\pm$ 1.0 & 73.4 $\pm$ 0.8 & 79.3 $\pm$ 1.3 & 79.2 $\pm$ 0.3 \\
RDM~\cite{nguyen2024domain} \hfill \tiny{(WACV '24)} & 88.4 $\pm$ 0.2 & 81.3 $\pm$ 1.6 & 97.1 $\pm$ 0.1 & 81.8 $\pm$ 1.1 & 87.2 $\pm$ 0.7 & 98.1 $\pm$ 0.2 & 64.9 $\pm$ 0.7 & 72.6 $\pm$ 0.5 & 77.9 $\pm$ 1.2 & 78.4 $\pm$ 0.4 \\
\rowcolor{gray!10}
RDM (reproduced) & 83.5 $\pm$ 0.7 & 81.5 $\pm$ 0.7 & 96.2 $\pm$ 0.5 & 79.9 $\pm$ 0.3 & 85.3 $\pm$ 0.3 & 95.9 $\pm$ 0.6 & 61.7 $\pm$ 0.8 & 67.2 $\pm$ 0.5 & 73.8 $\pm$ 1.0 & 74.6 $\pm$ 0.4 \\
URM~\cite{krishnamachari2024uniformly} \hfill \tiny{(TMLR '24)} & \small{N/A}\: $\pm$ \footnotesize{N/A} & \small{N/A}\: $\pm$ \footnotesize{N/A} & \small{N/A}\: $\pm$ \footnotesize{N/A} & \small{N/A}\: $\pm$ \footnotesize{N/A} & 87.2 $\pm$ 3.4 & \small{N/A}\: $\pm$ \footnotesize{N/A} & \small{N/A}\: $\pm$ \footnotesize{N/A} & \small{N/A}\: $\pm$ \footnotesize{N/A} & \small{N/A}\: $\pm$ \footnotesize{N/A} & 77.1 $\pm$ 0.2 \\
\rowcolor{gray!10}
URM (reproduced) & 83.5 $\pm$ 0.7 & 76.9 $\pm$ 1.7 & 94.7 $\pm$ 0.4 & 70.4 $\pm$ 2.0 & 81.4 $\pm$ 0.7 & 96.8 $\pm$ 0.6 & 64.5 $\pm$ 0.7 & 70.5 $\pm$ 0.2 & 76.4 $\pm$ 0.2 & 77.1 $\pm$ 0.2 \\
\rowcolor{gray!10}
CPCNet (Ours) & 86.0 $\pm$ 0.4 & 82.8 $\pm$ 0.9 & 95.4 $\pm$ 0.6 & 77.8 $\pm$ 0.7 & 85.5 $\pm$ 0.3 & 97.1 $\pm$ 0.3 & 63.2 $\pm$ 0.3 & 68.4 $\pm$ 1.8 & 74.9 $\pm$ 0.2 & 75.9 $\pm$ 0.5 \\
\rowcolor{gray!10}
PP-CPCNet (Ours) & 84.4 $\pm$ 0.1 & 80.8 $\pm$ 0.6 & 95.8 $\pm$ 0.4 & 79.2 $\pm$ 1.1 & 85.1 $\pm$ 0.3 & 97.1 $\pm$ 0.3 & 62.8 $\pm$ 1.8 & 67.4 $\pm$ 0.7 & 74.1 $\pm$ 1.6 & 75.3 $\pm$ 0.6 \\
\midrule
\multicolumn{11}{c}{\textbf{ViT-based Backbones}} \\
\midrule
SDViT-S~\cite{sultana2022self} \hfill \tiny{(ACCV '22)} & 87.6 $\pm$ 0.3 & 82.4 $\pm$ 0.4 & 98.0 $\pm$ 0.3 & 77.2 $\pm$ 1.0 & 86.3 $\pm$ 0.2 & 96.8 $\pm$ 0.5 & 64.2 $\pm$ 0.8 & \best{76.2}{0.4} & 78.5 $\pm$ 0.4 & 78.9 $\pm$ 0.4 \\
GMoE-S~\cite{lisparse} \hfill \tiny{(ICLR '23)} & \small{N/A}\: $\pm$ \footnotesize{N/A} & \small{N/A}\: $\pm$ \footnotesize{N/A} & \small{N/A}\: $\pm$ \footnotesize{N/A} & \small{N/A}\: $\pm$ \footnotesize{N/A} & 88.1 $\pm$ 0.1 & \small{N/A}\: $\pm$ \footnotesize{N/A} & \small{N/A}\: $\pm$ \footnotesize{N/A} & \small{N/A}\: $\pm$ \footnotesize{N/A} & \small{N/A}\: $\pm$ \footnotesize{N/A} & 80.2 $\pm$ 0.2 \\
GMoE-B~\cite{lisparse} \hfill \tiny{(ICLR '23)} & \small{N/A}\: $\pm$ \footnotesize{N/A} & \small{N/A}\: $\pm$ \footnotesize{N/A} & \small{N/A}\: $\pm$ \footnotesize{N/A} & \small{N/A}\: $\pm$ \footnotesize{N/A} & 89.4 $\pm$ 0.1 & \small{N/A}\: $\pm$ \footnotesize{N/A} & \small{N/A}\: $\pm$ \footnotesize{N/A} & \small{N/A}\: $\pm$ \footnotesize{N/A} & \small{N/A}\: $\pm$ \footnotesize{N/A} & \best{81.2}{0.1} \\
START-M-S~\cite{guo2024start} \hfill \tiny{(NeurIPS '24)} & 88.6 $\pm$ \scriptsize{N/A} & 83.2 $\pm$ \scriptsize{N/A} & 98.6 $\pm$ \scriptsize{N/A} & 77.8 $\pm$ \scriptsize{N/A} & 87.1 $\pm$ 0.3 & \small{N/A}\: $\pm$ \footnotesize{N/A} & \small{N/A}\: $\pm$ \footnotesize{N/A} & \small{N/A}\: $\pm$ \footnotesize{N/A} & \small{N/A}\: $\pm$ \footnotesize{N/A} & \small{N/A}\: $\pm$ \footnotesize{N/A} \\
START-M-B~\cite{guo2024start} \hfill \tiny{(NeurIPS '24)} & 88.7 $\pm$ \scriptsize{N/A} & 83.0 $\pm$ \scriptsize{N/A} & 98.5 $\pm$ \scriptsize{N/A} & 76.8 $\pm$ \scriptsize{N/A} & 86.8 $\pm$ 0.2 & \small{N/A}\: $\pm$ \footnotesize{N/A} & \small{N/A}\: $\pm$ \footnotesize{N/A} & \small{N/A}\: $\pm$ \footnotesize{N/A} & \small{N/A}\: $\pm$ \footnotesize{N/A} & \small{N/A}\: $\pm$ \footnotesize{N/A} \\
\rowcolor{gray!10}
CPCNet-S (Ours) & 87.6 $\pm$ 0.4 & 82.9 $\pm$ 0.7 & 96.9 $\pm$ 0.2 & 74.3 $\pm$ 1.5 & 85.4 $\pm$ 0.4 & 97.5 $\pm$ 0.3 & 64.3 $\pm$ 1.2 & 72.0 $\pm$ 0.4 & 79.6 $\pm$ 1.0 & 78.4 $\pm$ 0.4 \\
\rowcolor{gray!10}
PP-CPCNet-S (Ours) & 89.7 $\pm$ 0.7 & 84.7 $\pm$ 1.4 & 96.4 $\pm$ 0.4 & 78.9 $\pm$ 0.6 & 87.4 $\pm$ 0.4 & 97.1 $\pm$ 0.4 & 62.5 $\pm$ 0.5 & 73.3 $\pm$ 0.5 & 79.3 $\pm$ 0.9 & 78.0 $\pm$ 0.3 \\
\rowcolor{gray!10}
CPCNet-B (Ours) & \best{91.7}{0.6} & 85.3 $\pm$ 0.2 & \best{99.3}{0.4} & \best{82.1}{0.8} & \best{89.6}{0.3} & 97.9 $\pm$ 0.2 & \best{66.4}{1.0} & 73.6 $\pm$ 0.2 & \best{81.2}{1.4} & 79.8 $\pm$ 0.4 \\
\rowcolor{gray!10}
PP-CPCNet-B (Ours) & 91.4 $\pm$ 1.1 & \best{85.4}{0.7} & \best{99.3}{0.1} & \best{82.1}{0.6} & 89.5 $\pm$ 0.4 & \best{98.6}{0.6} & 65.5 $\pm$ 0.4 & 74.7 $\pm$ 0.2 & 79.3 $\pm$ 1.7 & 79.5 $\pm$ 0.5 \\
\midrule
\multicolumn{11}{c}{\textbf{SSM-based Backbones}} \\
\midrule
DGMamba-T~\cite{long2024dgmamba} \hfill \tiny{(ACM MM '24)} & 91.2 $\pm$ \scriptsize{N/A} & 86.9 $\pm$ \scriptsize{N/A} & 98.9 $\pm$ \scriptsize{N/A} & 87.0 $\pm$ \scriptsize{N/A} & 91.0 $\pm$ 0.1 & 97.7 $\pm$ \scriptsize{N/A} & 64.8 $\pm$ \scriptsize{N/A} & \best{79.3}{\scriptsize{N/A}} & 81.0 $\pm$ \scriptsize{N/A} & 80.7 $\pm$ 0.1 \\
DGMamba-S~\cite{long2024dgmamba} \hfill \tiny{(ACM MM '24)} & 94.1 $\pm$ \scriptsize{N/A} & 87.8 $\pm$ \scriptsize{N/A} & 99.6 $\pm$ \scriptsize{N/A} & \best{89.0}{\scriptsize{N/A}} & 92.6 $\pm$ \scriptsize{N/A} & \small{N/A}\: $\pm$ \footnotesize{N/A} & \small{N/A}\: $\pm$ \footnotesize{N/A} & \small{N/A}\: $\pm$ \footnotesize{N/A} & \small{N/A}\: $\pm$ \footnotesize{N/A} & \small{N/A}\: $\pm$ \footnotesize{N/A} \\
DGMamba-B~\cite{long2024dgmamba} \hfill \tiny{(ACM MM '24)} & \best{95.1}{\scriptsize{N/A}} & 89.2 $\pm$ \scriptsize{N/A} & \best{99.8}{\scriptsize{N/A}} & 87.9 $\pm$ \scriptsize{N/A} & \best{93.0}{\scriptsize{N/A}} & \small{N/A}\: $\pm$ \footnotesize{N/A} & \small{N/A}\: $\pm$ \footnotesize{N/A} & \small{N/A}\: $\pm$ \footnotesize{N/A} & \small{N/A}\: $\pm$ \footnotesize{N/A} & \small{N/A}\: $\pm$ \footnotesize{N/A} \\
START-M~\cite{guo2024start} \hfill \tiny{(NeurIPS '24)} & 93.3 $\pm$ \scriptsize{N/A} & 87.6 $\pm$ \scriptsize{N/A} & 99.1 $\pm$ \scriptsize{N/A} & 87.1 $\pm$ \scriptsize{N/A} & 91.8 $\pm$ 0.4 & \best{98.8}{\scriptsize{N/A}} & 67.0 $\pm$ \scriptsize{N/A} & 77.2 $\pm$ \scriptsize{N/A} & 82.3 $\pm$ \scriptsize{N/A} & \best{81.3}{0.3} \\
START-X~\cite{guo2024start} \hfill \tiny{(NeurIPS '24)} & 92.8 $\pm$ \scriptsize{N/A} & 87.4 $\pm$ \scriptsize{N/A} & 99.2 $\pm$ \scriptsize{N/A} & 87.5 $\pm$ \scriptsize{N/A} & 91.7 $\pm$ 0.5 & 98.7 $\pm$ \scriptsize{N/A} & 66.6 $\pm$ \scriptsize{N/A} & 77.0 $\pm$ \scriptsize{N/A} & \best{82.6}{\scriptsize{N/A}} & 81.2 $\pm$ 0.3 \\
\rowcolor{gray!10}
CPCNet-T (Ours) & 91.8 $\pm$ 0.7 & 86.3 $\pm$ 0.8 & 98.0 $\pm$ 0.4 & 82.3 $\pm$ 2.7 & 89.6 $\pm$ 0.7 & 97.5 $\pm$ 0.4 & 66.7 $\pm$ 0.4 & 72.9 $\pm$ 1.3 & 79.4 $\pm$ 1.3 & 79.1 $\pm$ 0.5 \\
\rowcolor{gray!10}
PP-CPCNet-T (Ours) & 90.5 $\pm$ 0.7 & 86.8 $\pm$ 1.2 & 97.6 $\pm$ 0.2 & 84.4 $\pm$ 0.4 & 89.8 $\pm$ 0.4 & 97.2 $\pm$ 0.4 & 66.4 $\pm$ 0.9 & 75.5 $\pm$ 1.1 & 78.6 $\pm$ 1.1 & 79.4 $\pm$ 0.4 \\
\rowcolor{gray!10}
CPCNet-S (Ours) & 93.6 $\pm$ 0.6 & 91.6 $\pm$ 0.4 & 99.5 $\pm$ 0.2 & 85.7 $\pm$ 1.4 & 92.6 $\pm$ 0.4 & 96.5 $\pm$ 0.6 & 65.8 $\pm$ 0.7 & 75.6 $\pm$ 1.4 & 82.4 $\pm$ 1.1 & 80.1 $\pm$ 0.5 \\
\rowcolor{gray!10}
PP-CPCNet-S (Ours) & 94.8 $\pm$ 0.4 & 89.9 $\pm$ 1.1 & 99.4 $\pm$ 0.2 & 85.8 $\pm$ 2.3 & 92.5 $\pm$ 0.6 & 97.3 $\pm$ 0.3 & 65.3 $\pm$ 0.7 & 75.3 $\pm$ 1.1 & 81.7 $\pm$ 0.3 & 79.9 $\pm$ 0.3 \\
\rowcolor{gray!10}
CPCNet-B (Ours) & 93.2 $\pm$ 0.4 & 87.8 $\pm$ 1.4 & 99.5 $\pm$ 0.1 & 84.4 $\pm$ 1.9 & 91.2 $\pm$ 0.6 & 96.5 $\pm$ 0.4 & 66.3 $\pm$ 1.3 & 75.3 $\pm$ 0.4 & 80.2 $\pm$ 0.5 & 79.6 $\pm$ 0.4 \\
\rowcolor{gray!10}
PP-CPCNet-B (Ours) & 93.6 $\pm$ 0.5 & \best{91.7}{0.1} & \best{99.8}{0.2} & 86.4 $\pm$ 0.7 & 92.9 $\pm$ 0.2 & 97.4 $\pm$ 0.2 & \best{67.9}{1.2} & 75.2 $\pm$ 0.6 & 80.5 $\pm$ 0.4 & 80.3 $\pm$ 0.4 \\
\bottomrule
\end{tabular}%
}
\end{table*}

\begin{table*}[t]
\centering
\caption{Detailed DG accuracies (\%) on \texttt{OfficeHome} and \texttt{TerraIncognita}. Best results within each backbone category are highlighted in \textbf{bold}. N/A denotes unreported results or standard deviations. Methods marked with $^\dagger$ are from the unified benchmark framework~\cite{gulrajanisearch}; other results are collected from the original papers or supplementary materials. Gray rows indicate results reproduced by us.}
\label{tab:officehome_terraincognita}
\resizebox{\linewidth}{!}{%
\begin{tabular}{p{4cm} | C C C C | c || C C C C | c }
\toprule
\multirow{2.5}{*}{\textbf{Method}} & \multicolumn{5}{c||}{\texttt{OfficeHome}} & \multicolumn{5}{c}{\texttt{TerraIncognita}} \\
\cmidrule(lr){2-6} \cmidrule(lr){7-11}
 & Art & Clipart & Product & Real & Avg.($\uparrow$) & L100 & L38 & L43 & L46 & Avg. ($\uparrow$) \\
\midrule
\multicolumn{11}{c}{\textbf{ResNet-50 Backbone}} \\
\midrule
ERM$^\dagger$~\cite{vapnik1998statistical} \hfill \tiny{(Book '98)} & 61.3 $\pm$ 0.7 & 52.4 $\pm$ 0.3 & 75.8 $\pm$ 0.1 & 76.6 $\pm$ 0.3 & 66.5 $\pm$ 0.3 & 49.8 $\pm$ 4.4 & 42.1 $\pm$ 1.4 & 56.9 $\pm$ 1.8 & 35.7 $\pm$ 3.9 & 46.1 $\pm$ 1.8 \\
\rowcolor{gray!10}
ERM (reproduced) & 63.2 $\pm$ 0.3 & 51.9 $\pm$ 0.6 & 75.3 $\pm$ 0.2 & 77.8 $\pm$ 0.9 & 67.1 $\pm$ 0.3 & 52.4 $\pm$ 1.8 & 47.4 $\pm$ 2.2 & 55.4 $\pm$ 1.5 & 40.8 $\pm$ 1.1 & 49.0 $\pm$ 0.8 \\
CORAL$^\dagger$~\cite{sun2016deep} \hfill \tiny{(ECCV '16)} & 65.3 $\pm$ 0.4 & 54.4 $\pm$ 0.5 & 76.5 $\pm$ 0.1 & 78.4 $\pm$ 0.5 & 68.7 $\pm$ 0.3 & 51.6 $\pm$ 2.4 & 42.2 $\pm$ 1.0 & 57.0 $\pm$ 1.0 & 39.8 $\pm$ 2.9 & 47.6 $\pm$ 1.0 \\
\rowcolor{gray!10}
CORAL (reproduced) & 64.0 $\pm$ 0.7 & 52.3 $\pm$ 0.6 & 75.8 $\pm$ 0.3 & 77.3 $\pm$ 0.2 & 67.3 $\pm$ 0.2 & 45.4 $\pm$ 2.7 & 35.4 $\pm$ 3.1 & 50.3 $\pm$ 1.4 & 35.4 $\pm$ 1.0 & 41.6 $\pm$ 1.1 \\
CDANN~\cite{li2018deep} \hfill \tiny{(ECCV '18)} & \small{N/A}\: $\pm$ \footnotesize{N/A} & \small{N/A}\: $\pm$ \footnotesize{N/A} & \small{N/A}\: $\pm$ \footnotesize{N/A} & \small{N/A}\: $\pm$ \footnotesize{N/A} & \small{N/A}\: $\pm$ \footnotesize{N/A} & \small{N/A}\: $\pm$ \footnotesize{N/A} & \small{N/A}\: $\pm$ \footnotesize{N/A} & \small{N/A}\: $\pm$ \footnotesize{N/A} & \small{N/A}\: $\pm$ \footnotesize{N/A} & \small{N/A}\: $\pm$ \footnotesize{N/A} \\
\rowcolor{gray!10}
CDANN (reproduced) & 64.3 $\pm$ 0.9 & 49.6 $\pm$ 0.4 & 75.2 $\pm$ 0.9 & 77.8 $\pm$ 0.3 & 66.7 $\pm$ 0.3 & 50.5 $\pm$ 2.1 & 47.8 $\pm$ 0.6 & 55.9 $\pm$ 1.0 & 41.5 $\pm$ 1.1 & 48.9 $\pm$ 0.7 \\
DGER~\cite{zhao2020domain} \hfill \tiny{(NeurIPS '20)} & \small{N/A}\: $\pm$ \footnotesize{N/A} & \small{N/A}\: $\pm$ \footnotesize{N/A} & \small{N/A}\: $\pm$ \footnotesize{N/A} & \small{N/A}\: $\pm$ \footnotesize{N/A} & \small{N/A}\: $\pm$ \footnotesize{N/A} & \small{N/A}\: $\pm$ \footnotesize{N/A} & \small{N/A}\: $\pm$ \footnotesize{N/A} & \small{N/A}\: $\pm$ \footnotesize{N/A} & \small{N/A}\: $\pm$ \footnotesize{N/A} & \small{N/A}\: $\pm$ \footnotesize{N/A} \\
SelfReg~\cite{kim2021selfreg} \hfill \tiny{(ICCV '21)} & 63.6 $\pm$ 1.4 & 53.1 $\pm$ 1.0 & 76.9 $\pm$ 0.4 & 78.1 $\pm$ 0.4 & 67.9 $\pm$ 0.7 & 48.8 $\pm$ 0.9 & 41.3 $\pm$ 1.8 & 57.3 $\pm$ 0.7 & 40.6 $\pm$ 0.9 & 47.0 $\pm$ 0.3 \\
\rowcolor{gray!10}
SelfReg (reproduced) & 65.6 $\pm$ 0.4 & 49.9 $\pm$ 0.3 & 75.7 $\pm$ 0.3 & 77.5 $\pm$ 0.5 & 67.2 $\pm$ 0.2 & 50.7 $\pm$ 3.1 & 46.0 $\pm$ 1.0 & 53.4 $\pm$ 0.9 & 38.8 $\pm$ 2.1 & 47.2 $\pm$ 1.0 \\
SagNet$^\dagger$~\cite{nam2021reducing} \hfill \tiny{(CVPR '21)} & 63.4 $\pm$ 0.2 & 54.8 $\pm$ 0.4 & 75.8 $\pm$ 0.4 & 78.3 $\pm$ 0.3 & 68.1 $\pm$ 0.1 & 53.0 $\pm$ 2.9 & 43.0 $\pm$ 2.5 & 57.9 $\pm$ 0.6 & 40.4 $\pm$ 1.3 & 48.6 $\pm$ 1.0 \\
\rowcolor{gray!10}
SagNet (reproduced) & 60.4 $\pm$ 1.4 & 49.1 $\pm$ 0.7 & 74.6 $\pm$ 0.1 & 75.0 $\pm$ 0.5 & 64.8 $\pm$ 0.4 & 49.1 $\pm$ 1.8 & 45.5 $\pm$ 0.5 & 56.5 $\pm$ 0.4 & 39.5 $\pm$ 0.3 & 47.7 $\pm$ 0.5 \\
SWAD~\cite{cha2021swad} \hfill \tiny{(NeurIPS '21)} & 66.1 $\pm$ 0.4 & 57.7 $\pm$ 0.4 & 78.4 $\pm$ 0.1 & 80.2 $\pm$ 0.2 & 70.6 $\pm$ 0.2 & 55.4 $\pm$ 0.0 & 44.9 $\pm$ 1.1 & 59.7 $\pm$ 0.4 & 39.9 $\pm$ 0.2 & 50.0 $\pm$ 0.3 \\
ARM$^\dagger$~\cite{zhang2021adaptive} \hfill \tiny{(NeurIPS '21)} & 63.9 $\pm$ 0.8 & 51.0 $\pm$ 0.7 & 74.5 $\pm$ 0.3 & 78.5 $\pm$ 0.6 & 67.0 $\pm$ 0.3 & 46.9 $\pm$ 1.2 & 41.9 $\pm$ 2.0 & 54.2 $\pm$ 0.2 & 38.8 $\pm$ 1.0 & 45.4 $\pm$ 0.6 \\
\rowcolor{gray!10}
ARM (reproduced) & 63.9 $\pm$ 0.8 & 51.0 $\pm$ 0.7 & 74.5 $\pm$ 0.3 & 78.5 $\pm$ 0.6 & 67.0 $\pm$ 0.3 & 46.9 $\pm$ 1.2 & 41.9 $\pm$ 2.0 & 54.2 $\pm$ 0.2 & 38.8 $\pm$ 1.0 & 45.4 $\pm$ 0.6 \\
IB-ERM~\cite{ahuja2021invariance} \hfill \tiny{(NeurIPS '21)} & \small{N/A}\: $\pm$ \footnotesize{N/A} & \small{N/A}\: $\pm$ \footnotesize{N/A} & \small{N/A}\: $\pm$ \footnotesize{N/A} & \small{N/A}\: $\pm$ \footnotesize{N/A} & \small{N/A}\: $\pm$ \footnotesize{N/A} & \small{N/A}\: $\pm$ \footnotesize{N/A} & \small{N/A}\: $\pm$ \footnotesize{N/A} & \small{N/A}\: $\pm$ \footnotesize{N/A} & \small{N/A}\: $\pm$ \footnotesize{N/A} & \best{56.4}{2.1} \\
\rowcolor{gray!10}
IB-ERM (reproduced) & 62.5 $\pm$ 0.7 & 51.7 $\pm$ 0.4 & 73.0 $\pm$ 0.5 & 75.3 $\pm$ 0.5 & 65.6 $\pm$ 0.3 & 56.5 $\pm$ 2.4 & 47.7 $\pm$ 1.1 & 58.5 $\pm$ 1.1 & 38.4 $\pm$ 0.3 & 50.3 $\pm$ 0.7 \\
IB-IRM~\cite{ahuja2021invariance} \hfill \tiny{(NeurIPS '21)} & \small{N/A}\: $\pm$ \footnotesize{N/A} & \small{N/A}\: $\pm$ \footnotesize{N/A} & \small{N/A}\: $\pm$ \footnotesize{N/A} & \small{N/A}\: $\pm$ \footnotesize{N/A} & \small{N/A}\: $\pm$ \footnotesize{N/A} & \small{N/A}\: $\pm$ \footnotesize{N/A} & \small{N/A}\: $\pm$ \footnotesize{N/A} & \small{N/A}\: $\pm$ \footnotesize{N/A} & \small{N/A}\: $\pm$ \footnotesize{N/A} & 54.1 $\pm$ 2.0 \\
\rowcolor{gray!10}
IB-IRM (reproduced) & 54.6 $\pm$ 0.5 & 46.9 $\pm$ 0.3 & 64.2 $\pm$ 0.3 & 66.6 $\pm$ 0.1 & 58.1 $\pm$ 0.2 & 45.3 $\pm$ 1.7 & 36.8 $\pm$ 2.9 & 51.7 $\pm$ 2.5 & 37.5 $\pm$ 0.8 & 42.8 $\pm$ 1.1 \\
Transfer~\cite{zhang2021quantifying} \hfill \tiny{(NeurIPS '21)} & \small{N/A}\: $\pm$ \footnotesize{N/A} & \small{N/A}\: $\pm$ \footnotesize{N/A} & \small{N/A}\: $\pm$ \footnotesize{N/A} & \small{N/A}\: $\pm$ \footnotesize{N/A} & \small{N/A}\: $\pm$ \footnotesize{N/A} & \small{N/A}\: $\pm$ \footnotesize{N/A} & \small{N/A}\: $\pm$ \footnotesize{N/A} & \small{N/A}\: $\pm$ \footnotesize{N/A} & \small{N/A}\: $\pm$ \footnotesize{N/A} & \small{N/A}\: $\pm$ \footnotesize{N/A} \\
\rowcolor{gray!10}
Transfer (reproduced) & 57.6 $\pm$ 1.0 & 46.4 $\pm$ 0.6 & 71.4 $\pm$ 0.2 & 74.7 $\pm$ 0.3 & 62.5 $\pm$ 0.3 & 41.0 $\pm$ 4.0 & 24.5 $\pm$ 3.4 & 48.7 $\pm$ 0.4 & 31.0 $\pm$ 1.2 & 36.3 $\pm$ 1.3 \\
EQRM~\cite{eastwood2022probable} \hfill \tiny{(NeurIPS '22)} & 60.5 $\pm$ 0.1 & 56.0 $\pm$ 0.2 & 76.1 $\pm$ 0.4 & 77.4 $\pm$ 0.3 & 67.5 $\pm$ 0.1 & 47.9 $\pm$ 1.9 & 45.2 $\pm$ 0.3 & 59.1 $\pm$ 0.3 & 38.8 $\pm$ 0.6 & 47.8 $\pm$ 0.6 \\
\rowcolor{gray!10}
EQRM (reproduced) & 64.7 $\pm$ 0.3 & 51.8 $\pm$ 0.4 & 75.6 $\pm$ 0.5 & 77.5 $\pm$ 0.3 & 67.4 $\pm$ 0.2 & 47.7 $\pm$ 1.5 & 38.9 $\pm$ 2.8 & 52.8 $\pm$ 0.2 & 40.7 $\pm$ 1.0 & 45.0 $\pm$ 0.8 \\
EoA~\cite{arpit2022ensemble} \hfill \tiny{(NeurIPS '22)} & \best{69.1}{\scriptsize{N/A}} & \best{59.8}{\scriptsize{N/A}} & 79.5 $\pm$ \scriptsize{N/A} & 81.5 $\pm$ \scriptsize{N/A} & \best{72.5}{\scriptsize{N/A}} & 57.8 $\pm$ \scriptsize{N/A} & 46.5 $\pm$ \scriptsize{N/A} & \best{61.3}{\scriptsize{N/A}} & 43.5 $\pm$ \scriptsize{N/A} & 52.3 $\pm$ \scriptsize{N/A} \\
ADRMX~\cite{demirel2023adrmx} \hfill \tiny{(arXiv '23)} & \small{N/A}\: $\pm$ \footnotesize{N/A} & \small{N/A}\: $\pm$ \footnotesize{N/A} & \small{N/A}\: $\pm$ \footnotesize{N/A} & \small{N/A}\: $\pm$ \footnotesize{N/A} & 68.3 $\pm$ \scriptsize{N/A} & \small{N/A}\: $\pm$ \footnotesize{N/A} & \small{N/A}\: $\pm$ \footnotesize{N/A} & \small{N/A}\: $\pm$ \footnotesize{N/A} & \small{N/A}\: $\pm$ \footnotesize{N/A} & 47.4 $\pm$ \scriptsize{N/A} \\
\rowcolor{gray!10}
ADRMX (reproduced) & 64.1 $\pm$ 0.4 & 50.6 $\pm$ 0.7 & 74.4 $\pm$ 0.3 & 77.5 $\pm$ 0.4 & 66.6 $\pm$ 0.2 & 47.0 $\pm$ 2.7 & 44.4 $\pm$ 1.2 & 55.1 $\pm$ 0.6 & 42.6 $\pm$ 0.9 & 47.2 $\pm$ 0.8 \\
MADG~\cite{dayal2023madg} \hfill \tiny{(NeurIPS '23)} & 68.6 $\pm$ 0.5 & 55.5 $\pm$ 0.2 & 79.6 $\pm$ 0.3 & 81.5 $\pm$ 0.4 & 71.3 $\pm$ 0.3 & 60.0 $\pm$ 1.2 & \best{51.8}{0.2} & 57.4 $\pm$ 0.3 & \best{45.6}{0.5} & 53.7 $\pm$ 0.5 \\
SAGM~\cite{wang2023sharpness} \hfill \tiny{(CVPR '23)} & 65.4 $\pm$ 0.4 & 57.0 $\pm$ 0.3 & 78.0 $\pm$ 0.3 & 80.0 $\pm$ 0.2 & 70.1 $\pm$ 0.2 & 54.8 $\pm$ 1.3 & 41.4 $\pm$ 0.8 & 57.7 $\pm$ 0.6 & 41.3 $\pm$ 0.4 & 48.8 $\pm$ 0.9 \\
GMDG~\cite{tan2024rethinking} \hfill \tiny{(CVPR '24)} & 68.9 $\pm$ 0.3 & 56.2 $\pm$ 1.7 & \best{79.9}{0.6} & \best{82.0}{0.4} & 70.7 $\pm$ 0.2 & \best{60.9}{2.5} & 47.3 $\pm$ 1.6 & 55.2 $\pm$ 0.5 & 41.0 $\pm$ 1.4 & 51.1 $\pm$ 0.9 \\
RDM~\cite{nguyen2024domain} \hfill \tiny{(WACV '24)} & 61.1 $\pm$ 0.4 & 55.1 $\pm$ 0.3 & 75.7 $\pm$ 0.5 & 77.3 $\pm$ 0.3 & 67.3 $\pm$ 0.4 & 52.9 $\pm$ 1.2 & 43.1 $\pm$ 1.0 & 58.1 $\pm$ 1.3 & 36.1 $\pm$ 2.9 & 47.5 $\pm$ 1.0 \\
\rowcolor{gray!10}
RDM (reproduced) & 65.5 $\pm$ 0.4 & 53.1 $\pm$ 0.3 & 75.8 $\pm$ 0.4 & 77.7 $\pm$ 0.6 & 68.0 $\pm$ 0.2 & 55.5 $\pm$ 1.9 & 45.4 $\pm$ 1.3 & 55.8 $\pm$ 0.9 & 41.3 $\pm$ 0.7 & 49.5 $\pm$ 0.6 \\
URM~\cite{krishnamachari2024uniformly} \hfill \tiny{(TMLR '24)} & \small{N/A}\: $\pm$ \footnotesize{N/A} & \small{N/A}\: $\pm$ \footnotesize{N/A} & \small{N/A}\: $\pm$ \footnotesize{N/A} & \small{N/A}\: $\pm$ \footnotesize{N/A} & 68.9 $\pm$ 0.6 & \small{N/A}\: $\pm$ \footnotesize{N/A} & \small{N/A}\: $\pm$ \footnotesize{N/A} & \small{N/A}\: $\pm$ \footnotesize{N/A} & \small{N/A}\: $\pm$ \footnotesize{N/A} & 49.3 $\pm$ 0.9 \\
\rowcolor{gray!10}
URM (reproduced) & 62.1 $\pm$ 1.1 & 50.2 $\pm$ 0.3 & 73.5 $\pm$ 0.4 & 75.2 $\pm$ 0.5 & 65.3 $\pm$ 0.3 & 54.4 $\pm$ 3.5 & 46.7 $\pm$ 0.7 & 57.3 $\pm$ 0.5 & 40.9 $\pm$ 0.6 & 49.8 $\pm$ 0.9 \\
\rowcolor{gray!10}
CPCNet (Ours) & 68.2 $\pm$ 0.4 & 54.2 $\pm$ 0.6 & 75.5 $\pm$ 0.4 & 79.2 $\pm$ 0.5 & 69.3 $\pm$ 0.3 & 44.3 $\pm$ 1.4 & 47.9 $\pm$ 0.4 & 57.2 $\pm$ 1.2 & 40.3 $\pm$ 2.8 & 47.4 $\pm$ 0.9 \\
\rowcolor{gray!10}
PP-CPCNet (Ours) & 67.8 $\pm$ 0.4 & 53.5 $\pm$ 0.3 & 75.9 $\pm$ 0.7 & 79.3 $\pm$ 0.2 & 69.1 $\pm$ 0.2 & 48.6 $\pm$ 2.3 & 44.5 $\pm$ 2.5 & 59.0 $\pm$ 0.6 & 37.8 $\pm$ 2.2 & 47.5 $\pm$ 1.0 \\
\midrule
\multicolumn{11}{c}{\textbf{ViT-based Backbones}} \\
\midrule
SDViT-S~\cite{sultana2022self} \hfill \tiny{(ACCV '22)} & 68.3 $\pm$ 0.8 & 56.3 $\pm$ 0.2 & 79.5 $\pm$ 0.3 & 81.8 $\pm$ 0.1 & 71.5 $\pm$ 0.2 & \best{55.9}{1.7} & 31.7 $\pm$ 2.6 & 52.2 $\pm$ 0.3 & 37.4 $\pm$ 0.6 & 44.3 $\pm$ 1.0 \\
GMoE-S~\cite{lisparse} \hfill \tiny{(ICLR '23)} & \small{N/A}\: $\pm$ \footnotesize{N/A} & \small{N/A}\: $\pm$ \footnotesize{N/A} & \small{N/A}\: $\pm$ \footnotesize{N/A} & \small{N/A}\: $\pm$ \footnotesize{N/A} & 74.2 $\pm$ 0.4 & \small{N/A}\: $\pm$ \footnotesize{N/A} & \small{N/A}\: $\pm$ \footnotesize{N/A} & \small{N/A}\: $\pm$ \footnotesize{N/A} & \small{N/A}\: $\pm$ \footnotesize{N/A} & 48.5 $\pm$ 0.4 \\
GMoE-B~\cite{lisparse} \hfill \tiny{(ICLR '23)} & \small{N/A}\: $\pm$ \footnotesize{N/A} & \small{N/A}\: $\pm$ \footnotesize{N/A} & \small{N/A}\: $\pm$ \footnotesize{N/A} & \small{N/A}\: $\pm$ \footnotesize{N/A} & 77.2 $\pm$ 0.4 & \small{N/A}\: $\pm$ \footnotesize{N/A} & \small{N/A}\: $\pm$ \footnotesize{N/A} & \small{N/A}\: $\pm$ \footnotesize{N/A} & \small{N/A}\: $\pm$ \footnotesize{N/A} & \best{49.3}{0.3} \\
START-M-S~\cite{guo2024start} \hfill \tiny{(NeurIPS '24)} & \small{N/A}\: $\pm$ \footnotesize{N/A} & \small{N/A}\: $\pm$ \footnotesize{N/A} & \small{N/A}\: $\pm$ \footnotesize{N/A} & \small{N/A}\: $\pm$ \footnotesize{N/A} & \small{N/A}\: $\pm$ \footnotesize{N/A} & \small{N/A}\: $\pm$ \footnotesize{N/A} & \small{N/A}\: $\pm$ \footnotesize{N/A} & \small{N/A}\: $\pm$ \footnotesize{N/A} & \small{N/A}\: $\pm$ \footnotesize{N/A} & \small{N/A}\: $\pm$ \footnotesize{N/A} \\
START-M-B~\cite{guo2024start} \hfill \tiny{(NeurIPS '24)} & \small{N/A}\: $\pm$ \footnotesize{N/A} & \small{N/A}\: $\pm$ \footnotesize{N/A} & \small{N/A}\: $\pm$ \footnotesize{N/A} & \small{N/A}\: $\pm$ \footnotesize{N/A} & \small{N/A}\: $\pm$ \footnotesize{N/A} & \small{N/A}\: $\pm$ \footnotesize{N/A} & \small{N/A}\: $\pm$ \footnotesize{N/A} & \small{N/A}\: $\pm$ \footnotesize{N/A} & \small{N/A}\: $\pm$ \footnotesize{N/A} & \small{N/A}\: $\pm$ \footnotesize{N/A} \\
\rowcolor{gray!10}
CPCNet-S (Ours) & 70.0 $\pm$ 1.5 & 56.1 $\pm$ 0.7 & 81.0 $\pm$ 0.6 & 84.1 $\pm$ 0.5 & 72.8 $\pm$ 0.4 & 52.3 $\pm$ 1.3 & \best{35.9}{3.0} & 52.2 $\pm$ 0.9 & 38.5 $\pm$ 0.6 & 44.7 $\pm$ 0.8 \\
\rowcolor{gray!10}
PP-CPCNet-S (Ours) & 71.2 $\pm$ 1.0 & 57.0 $\pm$ 0.3 & 80.9 $\pm$ 0.8 & 83.4 $\pm$ 0.6 & 73.1 $\pm$ 0.4 & 47.2 $\pm$ 1.9 & 34.1 $\pm$ 1.3 & 49.4 $\pm$ 1.1 & 39.0 $\pm$ 1.2 & 42.4 $\pm$ 0.7 \\
\rowcolor{gray!10}
CPCNet-B (Ours) & \best{77.9}{0.5} & 61.1 $\pm$ 0.4 & 83.2 $\pm$ 0.5 & 86.4 $\pm$ 0.3 & 77.1 $\pm$ 0.2 & 51.6 $\pm$ 0.4 & 33.2 $\pm$ 2.7 & 51.4 $\pm$ 0.5 & \best{42.4}{0.8} & 44.6 $\pm$ 0.7 \\
\rowcolor{gray!10}
PP-CPCNet-B (Ours) & \best{77.9}{0.4} & \best{62.4}{0.7} & \best{83.3}{0.2} & \best{86.9}{0.4} & \best{77.6}{0.2} & 49.8 $\pm$ 2.0 & 33.4 $\pm$ 2.2 & \best{54.7}{1.2} & 37.4 $\pm$ 0.8 & 43.8 $\pm$ 0.8 \\
\midrule
\multicolumn{11}{c}{\textbf{SSM-based Backbones}} \\
\midrule
DGMamba-T~\cite{long2024dgmamba} \hfill \tiny{(ACM MM '24)} & 75.6 $\pm$ \scriptsize{N/A} & 61.9 $\pm$ \scriptsize{N/A} & 83.8 $\pm$ \scriptsize{N/A} & 86.0 $\pm$ \scriptsize{N/A} & 76.8 $\pm$ 0.1 & 62.0 $\pm$ \scriptsize{N/A} & \best{67.7}{\scriptsize{N/A}} & 61.7 $\pm$ \scriptsize{N/A} & 46.9 $\pm$ \scriptsize{N/A} & 54.5 $\pm$ 0.1 \\
DGMamba-S~\cite{long2024dgmamba} \hfill \tiny{(ACM MM '24)} & \small{N/A}\: $\pm$ \footnotesize{N/A} & \small{N/A}\: $\pm$ \footnotesize{N/A} & \small{N/A}\: $\pm$ \footnotesize{N/A} & \small{N/A}\: $\pm$ \footnotesize{N/A} & \small{N/A}\: $\pm$ \footnotesize{N/A} & \small{N/A}\: $\pm$ \footnotesize{N/A} & \small{N/A}\: $\pm$ \footnotesize{N/A} & \small{N/A}\: $\pm$ \footnotesize{N/A} & \small{N/A}\: $\pm$ \footnotesize{N/A} & \small{N/A}\: $\pm$ \footnotesize{N/A} \\
DGMamba-B~\cite{long2024dgmamba} \hfill \tiny{(ACM MM '24)} & \small{N/A}\: $\pm$ \footnotesize{N/A} & \small{N/A}\: $\pm$ \footnotesize{N/A} & \small{N/A}\: $\pm$ \footnotesize{N/A} & \small{N/A}\: $\pm$ \footnotesize{N/A} & \small{N/A}\: $\pm$ \footnotesize{N/A} & \small{N/A}\: $\pm$ \footnotesize{N/A} & \small{N/A}\: $\pm$ \footnotesize{N/A} & \small{N/A}\: $\pm$ \footnotesize{N/A} & \small{N/A}\: $\pm$ \footnotesize{N/A} & \small{N/A}\: $\pm$ \footnotesize{N/A} \\
START-M~\cite{guo2024start} \hfill \tiny{(NeurIPS '24)} & 75.2 $\pm$ \scriptsize{N/A} & 62.0 $\pm$ \scriptsize{N/A} & \best{85.3}{\scriptsize{N/A}} & 85.8 $\pm$ \scriptsize{N/A} & 77.1 $\pm$ 0.2 & 70.1 $\pm$ \scriptsize{N/A} & 50.0 $\pm$ \scriptsize{N/A} & 63.0 $\pm$ \scriptsize{N/A} & 49.5 $\pm$ \scriptsize{N/A} & 58.2 $\pm$ 0.8 \\
START-X~\cite{guo2024start} \hfill \tiny{(NeurIPS '24)} & 75.5 $\pm$ \scriptsize{N/A} & \best{62.1}{\scriptsize{N/A}} & 85.2 $\pm$ \scriptsize{N/A} & 85.5 $\pm$ \scriptsize{N/A} & 77.1 $\pm$ 0.1 & \best{70.7}{\scriptsize{N/A}} & 49.5 $\pm$ \scriptsize{N/A} & 64.0 $\pm$ \scriptsize{N/A} & 49.0 $\pm$ \scriptsize{N/A} & \best{58.3}{0.8} \\
\rowcolor{gray!10}
CPCNet-T (Ours) & 69.0 $\pm$ 0.4 & 54.6 $\pm$ 1.5 & 77.5 $\pm$ 0.3 & 79.8 $\pm$ 0.2 & 70.2 $\pm$ 0.4 & 60.3 $\pm$ 0.3 & 46.5 $\pm$ 0.8 & 61.9 $\pm$ 0.9 & 47.6 $\pm$ 1.2 & 54.1 $\pm$ 0.4 \\
\rowcolor{gray!10}
PP-CPCNet-T (Ours) & 68.0 $\pm$ 1.0 & 52.5 $\pm$ 1.3 & 76.8 $\pm$ 0.8 & 79.2 $\pm$ 0.6 & 69.1 $\pm$ 0.5 & 57.3 $\pm$ 4.6 & 47.1 $\pm$ 2.6 & 62.2 $\pm$ 1.4 & 48.9 $\pm$ 2.8 & 53.9 $\pm$ 1.5 \\
\rowcolor{gray!10}
CPCNet-S (Ours) & 77.1 $\pm$ 0.7 & 58.6 $\pm$ 0.2 & 81.5 $\pm$ 0.1 & 84.3 $\pm$ 0.5 & 75.4 $\pm$ 0.2 & 59.7 $\pm$ 0.5 & 51.3 $\pm$ 0.3 & \best{64.8}{0.3} & 49.9 $\pm$ 1.1 & 56.4 $\pm$ 0.3 \\
\rowcolor{gray!10}
PP-CPCNet-S (Ours) & 76.2 $\pm$ 1.4 & 59.0 $\pm$ 0.8 & 82.7 $\pm$ 0.8 & 84.7 $\pm$ 0.4 & 75.6 $\pm$ 0.5 & 64.5 $\pm$ 0.9 & 47.9 $\pm$ 1.3 & 63.5 $\pm$ 0.2 & 51.1 $\pm$ 1.3 & 56.8 $\pm$ 0.5 \\
\rowcolor{gray!10}
CPCNet-B (Ours) & \best{79.7}{0.8} & 61.6 $\pm$ 0.5 & 85.2 $\pm$ 0.3 & 85.8 $\pm$ 0.5 & \best{78.1}{0.3} & 66.4 $\pm$ 2.0 & 49.9 $\pm$ 1.4 & 64.3 $\pm$ 0.9 & \best{50.1}{1.9} & 57.7 $\pm$ 0.8 \\
\rowcolor{gray!10}
PP-CPCNet-B (Ours) & 79.3 $\pm$ 0.5 & 61.5 $\pm$ 1.0 & 84.4 $\pm$ 0.3 & \best{86.1}{0.3} & 77.8 $\pm$ 0.3 & 65.5 $\pm$ 1.5 & 52.5 $\pm$ 2.8 & 63.1 $\pm$ 0.5 & 49.6 $\pm$ 1.3 & 57.7 $\pm$ 0.9 \\
\bottomrule
\end{tabular}%
}
\end{table*}

\clearpage

\begin{table}[t!]
    \centering
    
    \caption{Hyperparameter search over projection dimension ($d$) and unfolding stages ($T$) with a ResNet-50 backbone. Results are averaged across four datasets. The selected configuration ($d=256, T=3$) is used for all subsequent experiments.}
    \label{tab:cpcnet_ablation}
    \resizebox{\linewidth}{!}{%
    \begin{tabular}{c | c c c c c c}
    \toprule
    \multirow{2.5}{*}{\makecell{\textbf{Proj.} \\ \textbf{Dim.} (\(d\))}} & \multicolumn{6}{c}{\textbf{Unfolding Stages} ($T$)} \\
    \cmidrule(lr){2-7}
     & \textbf{$T=1$} & \textbf{$T=2$} & \textbf{$T=3$} & \textbf{$T=4$} & \textbf{$T=5$} & \textbf{$T=6$} \\
    \midrule
    64  & 68.5 $\pm$ 0.2 & 68.9 $\pm$ 0.4 & 68.0 $\pm$ 0.3 & 68.7 $\pm$ 0.2 & 69.1 $\pm$ 0.6 & 68.7 $\pm$ 0.4 \\
    128 & 68.6 $\pm$ 0.1 & 68.5 $\pm$ 0.2 & 69.2 $\pm$ 0.6 & 68.6 $\pm$ 0.6 & 68.6 $\pm$ 0.2 & 68.8 $\pm$ 0.2 \\
    256 & 68.9 $\pm$ 0.2 & 68.8 $\pm$ 0.4 & \textbf{69.5} $\pm$ 0.5 & 68.8 $\pm$ 0.2 & 69.0 $\pm$ 0.2 & 69.1 $\pm$ 0.3 \\
    512 & 68.8 $\pm$ 0.5 & 68.8 $\pm$ 0.2 & 68.7 $\pm$ 0.1 & 68.9 $\pm$ 0.3 & 69.2 $\pm$ 0.1 & 68.4 $\pm$ 0.2 \\
    \bottomrule
    \end{tabular}%
    }
    
    \vspace{1.5em} 
    
    \caption{Ablation on the initial projection dimension $d_1$ with the progressive cascade depth fixed to $T=1$.}
    \label{tab:ablation_d1}
    \resizebox{\linewidth}{!}{%
    \begin{tabular}{c | c}
    \toprule
    \textbf{Initial Dim.} ($d_1$) & \textbf{Progressive Cascade Depths} ($T=1$) \\
    \midrule
    32  & 69.0 $\pm$ 0.3 \\
    64  & 68.9 $\pm$ 0.2 \\
    128 & \textbf{69.2} $\pm$ 0.3 \\
    256 & 68.8 $\pm$ 0.2 \\
    512 & 68.7 $\pm$ 0.0 \\
    \bottomrule
    \end{tabular}%
    }
    
    \vspace{1.5em} 
    
    \caption{Ablation on the progressive cascade depth $T$ with the initial projection dimension fixed to the optimal $d_1=128$.}
    \label{tab:ppcpcnet_ablation}
    \resizebox{\linewidth}{!}{%
    \begin{tabular}{c | c c c c c c}
    \toprule
    \multirow{2.5}{*}{\makecell{\textbf{Initial} \\ \textbf{Dim.} ($d_1$)}} & \multicolumn{6}{c}{\textbf{Progressive Cascade Depths} ($T$)} \\
    \cmidrule(lr){2-7}
     & \textbf{$T=1$} & \textbf{$T=2$} & \textbf{$T=3$} & \textbf{$T=4$} & \textbf{$T=5$} & \textbf{$T=6$} \\
    \midrule
    128 & \textbf{69.2} $\pm$ 0.3 & 68.3 $\pm$ 0.4 & 68.8 $\pm$ 0.2 & 68.3 $\pm$ 0.1 & 68.5 $\pm$ 0.4 & 68.6 $\pm$ 0.3 \\
    \bottomrule
    \end{tabular}%
    }
\end{table}


\end{document}